%% file: manuscript.tex
\documentclass{article}
\usepackage[table]{xcolor}
\usepackage{iclr2027_conference,times}
\usepackage[T1]{fontenc}
\usepackage[utf8]{inputenc}
\usepackage{inconsolata}
\usepackage{amsmath,amssymb,booktabs,array,graphicx,float,placeins}
\floatstyle{ruled}
\newfloat{algorithm}{tbp}{loa}
\floatname{algorithm}{Algorithm}
\floatstyle{plain}
\usepackage{hyperref,url}
\input{results_tables_style.tex}

\input{paperB_main_tables_style_addendum.tex}
\input{policy_listings_style.tex}
\hypersetup{hidelinks,pdfauthor={},pdftitle={From Answers to Policies: Efficient In-Context Learning System through Emulating Expert Investigation}}
\providecommand{\tightlist}{\setlength{\itemsep}{0pt}\setlength{\parskip}{0pt}}

\title{\raggedright\hyphenpenalty=10000 From Answers to Policies: Efficient In-Context Learning System through Emulating Expert Investigation}
\input{author_block.tex}
\iclrfinalcopy
\date{September 28, 2026}
\begin{document}
\maketitle
\lhead{Preprint}
\raggedbottom
\brokenpenalty=10000
\input{body.tex}
\end{document}

%% file: results_tables_style.tex
\definecolor{RTpihf}{HTML}{E8F1F8}
\definecolor{RTmuted}{gray}{0.45}

\providecommand{\RTfont}{\fontsize{9.5}{11.5}\selectfont}
\providecommand{\RTsetup}{\RTfont\setlength{\tabcolsep}{5pt}\renewcommand{\arraystretch}{1.15}}

\providecommand{\RTgroup}[2]{\multicolumn{#1}{@{}l}{\textit{#2}}}
\providecommand{\RTin}{\hspace{0.7em}}

\providecommand{\RTmark}[1]{\rlap{\textsuperscript{\,#1}}}

\providecommand{\RTref}[1]{\textcolor{RTmuted}{#1}}

\providecommand{\RTnr}{\textcolor{RTmuted}{n.r.}}

%% file: paperB_main_tables_style_addendum.tex
\definecolor{RTaccent}{HTML}{1F5A96}
\providecommand{\RTgain}[1]{\textcolor{RTaccent}{\textbf{#1}}}

\providecommand{\RTpanel}[2]{\multicolumn{#1}{@{}l}{\textsc{#2}}}

%% file: policy_listings_style.tex
\usepackage[breakable,skins]{tcolorbox}
\usepackage{fvextra}
\usepackage{needspace}
\usepackage{newunicodechar}
\newunicodechar{←}{\ensuremath{\leftarrow}}
\definecolor{policylistingbg}{HTML}{F2F3F5}
\definecolor{policylistinglabel}{HTML}{4A5563}
\newcounter{policylisting}
\newlength{\policyindent}
\newcommand{\policylistingcaption}[1]{%
  \par\needspace{12\baselineskip}\medskip\noindent
  \refstepcounter{policylisting}%
  {\small\textbf{Listing \thepolicylisting}\enspace$\vert$\enspace #1}\par
  \vspace{3pt}\nointerlineskip\noindent\rule{\linewidth}{0.4pt}\par\nopagebreak\vspace{5pt}\nopagebreak[4]}
\newtcolorbox{policybox}{breakable, enhanced, colback=policylistingbg, frame hidden,
  sharp corners, boxrule=0pt, left=7pt, right=7pt, top=6pt, bottom=6pt,
  before skip=0pt, after skip=14pt}
\newcommand{\policysection}[1]{%
  \par\smallskip\noindent{\sffamily\scriptsize\bfseries\color{policylistinglabel}#1}\par\nopagebreak\vspace{1pt}}
\DefineVerbatimEnvironment{PolicyVerbatim}{Verbatim}{breaklines, fontsize=\footnotesize,
  fontfamily=zi4, breaksymbolleft={}, breakindent=0pt}

%% file: author_block.tex
\author{\begin{minipage}{0.95\textwidth}\raggedright
Minh-Ha Nguyen\textsuperscript{1,*}, Ngoc-Ngo Quang Tran\textsuperscript{1},\\
Thuy Dung Nguyen\textsuperscript{1}, Cathy Shyr\textsuperscript{2,*}\\[0.5em]
\normalfont\small
\textsuperscript{1}Vanderbilt University\\
\textsuperscript{2}Vanderbilt University Medical Center\\[0.5em]
\textsuperscript{*}Corresponding authors: Minh-Ha Nguyen (\href{mailto:ha.m.nguyen@vanderbilt.edu}{ha.m.nguyen@vanderbilt.edu})\\
and Cathy Shyr (\href{mailto:cathy.shyr@vumc.org}{cathy.shyr@vumc.org}).
\end{minipage}}
\hypersetup{pdfauthor={Minh-Ha Nguyen, Ngoc-Ngo Quang Tran, Thuy Dung Nguyen, Cathy Shyr}}

%% file: body.tex
\begin{abstract}
Pretrained large language models offer a practical foundation for learning useful behavior from few task-specific examples. We argue that current prompt and context optimization methods underuse the extensive knowledge and reasoning capabilities of trillion-parameter models. These capabilities can make adaptation more sample-efficient, more compute efficient and at no performance loss when organized around how human experts investigate failures. We formalize Policy Iteration with Human Feedback (PIHF), which makes this implicit procedure explicit for LLM agents to execute, and build its automated implementation, PIHF-MCP. Initialized from clinician feedback on rare-disease diagnosis, PIHF-MCP supplies the expert procedure, testing tools, review and persistent inquiry records to develop reusable task policies. Across general reasoning benchmarks (BIG-Bench Extra Hard, HoVer and LiveBench-Math), PIHF-MCP improved performance of the baseline model by 16.9, 22.2 and 4.7 percentage points, respectively. With a matched baseline model, development used about 1/5 of the labelled examples and 4\% of the task rollouts reported by a previous SOTA in-context optimizer, making it about 9 times faster and 3 times cheaper at comparable or higher scores. In a low-data rare-disease diagnosis setting, policies developed from previous SOTA prompt optimizers trailed a previously published PIHF-developed system on every held-out cohort (on average 16 percentage points). These findings support a route to more efficient inference-time scaling: PIHF-MCP develops reusable policies from a few examples that improve performance on unseen cases and across models. Because each policy comes from an explicit, recorded investigation, the process also keeps humans in the loop and enables ownership and learning, making it well suited to high-stakes decisions.
\end{abstract}

\section{Introduction}\label{introduction}

Learning useful behavior from few task-specific examples remains a central objective in machine learning. Pretrained large language models made this idea broadly deployable across heterogeneous real-world tasks: a single fixed model can adapt through instructions and demonstrations across many tasks without weight updates \citep{brown_language_2020}. These capabilities, together with the development of pretrained LLMs in the trillion-parameter range \citep{kimi_team_kimi_2026, deepseek-ai_deepseek-v4_2026}, motivate using prior model learning as a foundation for adaptation when task-specific examples are scarce.

The ability of a machine to adapt quickly to new situations from limited examples is especially consequential in rare-disease diagnosis, where clinical heterogeneity and limited clinician familiarity contribute to prolonged diagnostic odysseys \citep{marwaha_guide_2022, faye_time_2024}. A useful learning system should therefore capture limited but precious case information and expert input, then carry the resulting lessons to unseen diseases, genes and atypical presentations. Prior work on rare-disease diagnosis showed that expert investigation of diagnostic failures can turn a small set of development cases into a reusable task policy for LLM to execute \citep{nguyen_teaching_2026}. Here, a policy is a set of instructions that specifies how an LLM should reason and use available tools to solve a task.

We formalize \textbf{Policy Iteration with Human Feedback (PIHF)} and introduce its automated implementation, \textbf{PIHF-MCP}. PIHF makes implicit expert investigative procedures explicit. LLM agents examine failures, tests competing explanations and develops revisions within an environment that supplies execution evidence, review and persistent inquiry records. The resulting policy can guide subsequent task execution without repeating the investigation, allowing computation spent on a few development cases to benefit many later cases, including those handled by a different executor.

Our evaluation begins by testing published prompt optimizers in the low-data clinical setting established by the previously clinical study \citep{nguyen_teaching_2026}. We then use general reasoning benchmarks to test whether PIHF-MCP can automate policy improvement beyond medicine and produce policies that transfer across test cases and executors.

We make three contributions:

\begin{enumerate}
\def\labelenumi{\arabic{enumi}.}
\tightlist
\item
  \textbf{An executable system for expert-informed investigation.} An environment allows LLM to autonomously emulate the implicit procedural failure investigation of human experts: review failure traces, generating competing hypotheses, targeted probes to localize the effect, ultimately enable end-to-end hypothesis testing.
\item
  \textbf{Evaluation generalizability on various fields} We showed that PIHF-MCP can develop policies that consistently improve performance on held-out test sets on general reasoning benchmarks. Compared to GEPA, PIHF-MCP development is about 9 times faster and 3 times cheaper, at 1/5 required sample size and offer comparable or higher performance gain.
\item
  \textbf{Characterization of investigation dynamics.} We observed interesting ``duality of investigation dynamics'' such as adaptively switching from broad architecture search to localized optimization when evidence changes, persistence in repeated error-diagnosis but also stopping early when the LLM investigator deemed further work would lead to overfit; and increased discrimination quality of generating hypotheses with more accumulating failures.
\end{enumerate}

\section{Background and motivation}\label{background-and-motivation}

Pretrained LLMs provide a practical foundation for learning reusable procedures from limited task experience: instructions and examples can change behavior across tasks without updating model weights \citep{radford_language_2019, brown_language_2020}. Later approaches in the literature extend this capability through reflective prompt evolution and accumulate strategies in persistent contextual playbooks \citep{agrawal_gepa_2026, zhang_agentic_2026}. These methods establish that task experience can improve an external artifact used on later cases. However, these optimization processes select revisions by task scores alone and operate largely as a black box that is difficult for human to understand, monitor or intervene in.

This opacity matters during high-stakes scenarios where human users are legally responsible for their decisions. Systems optimized on outcomes can reach the right answers for the wrong reasons, which limits their generalizability and real-world deployment. For example, in clinical medicine, physicians reviewing GPT-4V responses to NEJM Image Challenges found flawed rationales even when the final choice was correct \citep{jin_hidden_2024}, and pneumonia classifiers exploited hospital-associated cues, with performance gains that did not consistently transfer to external hospitals \citep{zech_variable_2018}.

Learning from few cases therefore requires \textbf{judgments} about which lessons remain clinically sound in unfamiliar presentations, objectives that are hard to specify mathematically. Human judgment can supply such objectives. Preference-based reinforcement learning showed how human judgments can specify objectives that are difficult to express as programmatic rewards \citep{christiano_deep_2017}, and RLHF extended this approach to instruction following \citep{ouyang_training_2022}. In addition, a previous publication has shown that adapting baseline LLMs with Policy Iteration with Human Feedback (PIHF) procedure on a 50-sample development set of rare diseases created a policy that generalizes to 1,193 held-out cases (679 diseases absent from development) and transfers across model backbones \citep{nguyen_teaching_2026}.

These developments motivate to explore whether expert-guided process can be \textbf{generalized and automated} into an optimization procedure that learns from few examples, as human experts do, costs less to run and maintains performance compared to previous prompt optimizer. The procedure should also remain understandable: it keeps humans in the loop and makes its rationale easy to communicate to decision makers, without slowing the agent. These properties matter most in high-stakes settings where human experts remain responsible for decisions and ownership of the improvement process lets them direct and strengthen it.

\section{System overview and methods}\label{system-overview-and-methods}

PIHF-MCP operationalizes ``expert failure diagnosis'' into a sequential environment that an agent can carry out: inspect task execution, compare explanations, test revisions and decide which policy to retain (Figure \ref{display:framework}A--B; detailed expert-practice mapping in Figure \ref{display:expert_system}).

The system comprises five agents: a Causal Investigator (CI), a Curator, a Critic, a Verifier and an Executor. Within one complete cycle, the CI owns and uses other actors for end-to-end hypothesis testing. CI directs task executors apply the current policy to individual cases, generating the execution traces and result panel for the CI to review. The CI can call on the Critic and Verifier to help check the strength of the analysis and the ``generalizability radius'' of the proposal to assess whether a proposed repair could apply beyond the observed failure. A Python controller schedules jobs and coordinates the agents' actions.

Clinician feedback from rare-disease policy development supplied the initial investigative practices, including examining evidence use, preserving plausible alternatives and testing whether a proposed repair should be reused \citep{nguyen_teaching_2026}. PIHF-MCP implements these practices through prompts for actors, observations, tests and review procedures used throughout the investigation.

\setcounter{figure}{0}
\begin{figure}[H]
\centering
\includegraphics[width=\linewidth]{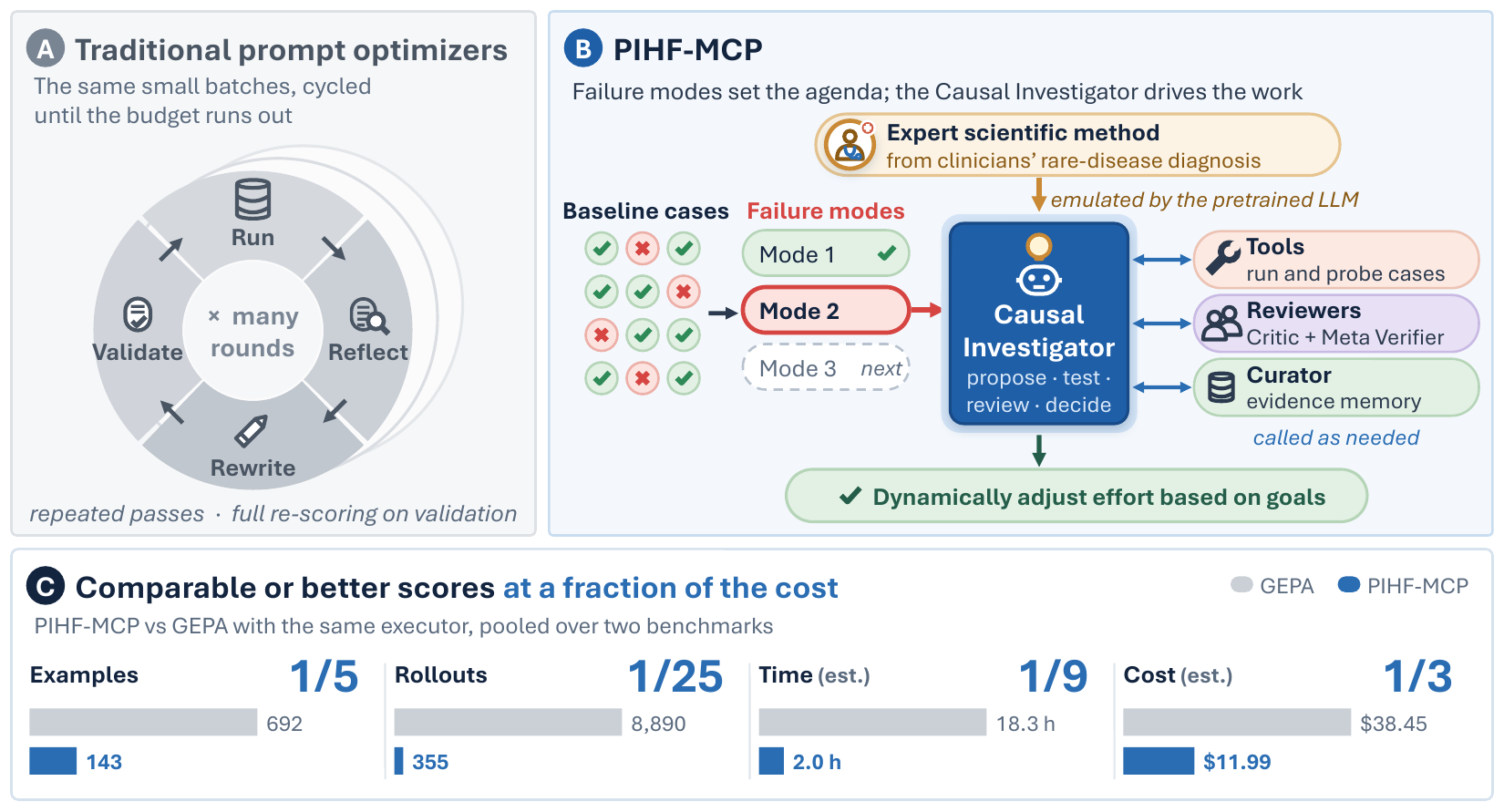}
\caption{PIHF-MCP turns expert failure investigation into executable learning loops. \textbf{A}, Conventional prompt-optimization rounds repeatedly run, reflect, rewrite and validate. \textbf{B}, PIHF-MCP organizes inquiry around observed failure modes; the Causal Investigator selects tests and revisions, drawing on tools, reviewers and curator memory within the task contract. Clinician-derived investigative practice informs its guidance. The Meta Verifier is the Verifier role described in Methods.}\label{display:framework}
\end{figure}

\subsection{Human specification and delegated investigation}\label{human-specification-and-delegated-investigation}

We tested PIHF-MCP on three general reasoning benchmarks: BIG-Bench Extra Hard (BBEH) \citep{kazemi_big-bench_2025}, HoVer \citep{jiang_hover_2020} and LiveBench-Math \citep{white_livebench_2025}.

The PIHF-MCP optimization process starts with a human specifying the task objective and which parts of the policy or program may change. The specification also defines the available development and adaptive-validation data, the protected test set, the scoring procedure and model assignments, resource limits, review roles and stopping conditions, including the compute budget. These choices determine what the CI can investigate and revise.

Within this specification, the CI drives the workflow by examining evidence, proposing changes and deciding on candidate-selection. These decisions operated within a declared revision surface for each benchmark: HoVer's modular program, LiveBench-Math's solve instruction, and BBEH's reasoning instruction.

\subsection{Policy representation and the investigation loop}\label{policy-representation-and-the-investigation-loop}

Let \(A=(P,T)\) comprise the natural-language task policy \(P\) and available executor tools \(T\). For a case \(z\sim\mathcal D\), executor model \(M\), held fixed within a comparison, produces a trajectory \(\tau\sim\pi_{M,A}(\cdot\mid z)\) with final answer \(\widehat y(\tau)\), scored by the task evaluator \(R(z,\widehat y)\). PIHF seeks an artifact within the human-authorized space \(\mathcal A_\Gamma\) that improves expected task performance:

\[
\max_{A\in\mathcal A_\Gamma} J(A),\qquad
J(A)=\mathbb E_{z\sim\mathcal D,\;\tau\sim\pi_{M,A}(\cdot\mid z)}
\bigl[R(z,\widehat y(\tau))\bigr].
\]

The investigation state \(\mathcal X_t=(M,A_t,C_t,S_t;\Gamma)\) includes CI guidance \(C_t\), the inquiry record \(S_t\) and the contract \(\Gamma\) governing evidence access, revisions, resources and decision roles. Algorithm \ref{alg:pihf} pursues the objective through bounded investigation and comparison on declared development and adaptive-validation panels. For a frozen revision \(A'_t\), incumbent and candidate evidence \(E_t,E'_t\) inform \(d_t=\operatorname{Admit}_\Gamma(E_t,E'_t,A_t,A'_t)\): promotion adopts \(A'_t\), retention preserves \(A_t\), and revision or rollback follows the contract. Alternating evaluation and artifact improvement parallels generalized policy iteration \citep{sutton_reinforcement_2018}.

\begin{algorithm}[H]
\caption{PIHF-MCP: investigation and policy iteration}\label{alg:pihf}
\small
\textbf{Input:} $M,A_0,C_0,S_0,\Gamma$; declared development and adaptive-validation data.
\begin{enumerate}
\setlength{\itemsep}{0pt}\setlength{\parskip}{0pt}\setlength{\parsep}{0pt}
\item Establish baseline traces and outcomes for $A_0$ under $\Gamma$.
\item \textbf{While} resources remain and stopping conditions are unmet:
\item \quad Inspect failures; record rival explanations and distinguishing predictions in $S_t$.
\item \quad Run permitted probes; update explanations and choose a local or structural repair.
\item \quad Freeze $A'_t$; obtain required reviews and compare it with $A_t$ under $\Gamma$.
\item \quad Apply $\operatorname{Admit}_\Gamma$; record the decision and curate evidence for continuation.
\item \textbf{End while}
\item Freeze the selected artifact and evaluate it on protected cases.
\end{enumerate}
\textbf{Outer loop:} Human corrections update the inquiry; authorized distillation refines $C$ for later inquiries. Contract changes require a new invocation.
\end{algorithm}

\subsection{Investigating failures with layered feedback}\label{investigating-failures-with-layered-feedback}

The CI begins by inspecting the starting policy or program and its available baseline behavior. Execution traces expose intermediate outputs and tool results alongside the final outcome. The CI uses these observations and its task knowledge to formulate competing explanations. In document retrieval, for example, a required title may never have been retrieved, may have been lost during an intermediate summary, or may not have informed the relation targeted by a later query. The CI inspects the retrieved passages and downstream queries to determine which explanations remain plausible. It records these alternatives and the observations that would distinguish them, then chooses a test that could change its next decision.

Targeted tests connect those explanations to observable effects. A revision that preserves information from a retrieved passage can be examined for its effect on a later query as well as its final retrieval score. This local evidence can support further investigation even when another failure still prevents the case from succeeding. Conversely, a favorable final result does not by itself establish that the proposed repair worked through its intended mechanism. Trace inspection, local checks, and case probes thus answer different questions as the CI develops and tests a candidate revision.

\subsection{Reviewing, retaining and reusing revisions}\label{reviewing-retaining-and-reusing-revisions}

A promising local result leads to evaluation on a broader development panel. Critics examine proposed explanations and candidate comparisons, while verifiers review assigned evaluation plans and results. End-to-end comparisons determine whether a revision improves the task across the development cases. Adaptive validation supplies additional evidence for selection; the CI then makes the delegated decision to retain, revise or reject the candidate.

The inquiry record preserves the evidence behind these decisions: hypotheses, predictions, test results, candidate versions and unresolved alternatives. The Curator maintains summaries linked to the underlying records so that later steps can recover what was tested and why a candidate was retained or rejected. This memory supports continued investigation and stopping decisions within the declared resource limits.

After selection, the final task artifact is frozen and evaluated on protected cases excluded from adaptation. Reusing the investigation procedure on a different task produces a new task-specific artifact, allowing the same PIHF system to be evaluated across clinical diagnosis and general reasoning benchmarks.

\section{Evaluation and results}\label{evaluation-and-results}

\subsection{Evaluation design}\label{evaluation-design}

The evaluation has two parts: we tested previous SOTA prompt optimizers under the common low-clinical-data, followed by automated PIHF-MCP development on general reasoning benchmarks. We used the standard development/validation and test split. Rollout count as one task execution is one returned executor trajectory on one case. For the admitted policy, we ran the same policy three times on the test set to estimate execution variability and reporting the mean and standard deviation.

The automated evaluation used a Luna-low executor for the primary peformance gain study. Separate GPT-4.1-mini inquiries provided the matched-backbone HoVer and LiveBench-Math comparisons with published GEPA results \citep{agrawal_gepa_2026}.

\subsection{Significant clinical gains extend beyond the development benchmark}\label{significant-clinical-gains-extend-beyond-the-development-benchmark}

We adopted the development cases and held-out cohorts of a previously published rare-disease study, in which a PIHF-developed system, liteOdyssey, was built from 50 LIRICAL cases spanning 43 diseases \citep{nguyen_teaching_2026}. Evaluation used 320 held-out LIRICAL cases and 873 PhenoPacket Store cases. The portfolio covered 722 diseases, 679 absent from development. More than half of these benchmarks cases are ultra-rare diseases, define as prevalent less than 1 per million.

On the same pool of 50 clinical cases and GPT-5.4 executor, we used GEPA and ACE \citep{agrawal_gepa_2026, zhang_agentic_2026}. Each method could use up to 2,100 returned diagnostic trajectories, the maximum cap used in the published PIHF procedure, and ended when it reached that ceiling or its own endpoint, whichever came first. This comparison tests what each system could develop under a common diagnostic-execution allowance and the same access to biomedical information sources.

\begin{table}[H]
\centering
\caption{PIHF, GEPA and ACE policies developed from the same 50 clinical cases.}\label{display:clinical_comparison}
\input{tables/clinical_comparison.tex}
\par\smallskip
\begin{minipage}{\linewidth}\footnotesize
Disease recall (\%) uses exact OMIM identifiers; R@1 and R@5 denote recall within the first and first five ranked diagnoses. Cohort sizes appear in headers. Bold, shaded cells identify the best result in each column. Development results are descriptive. Superscript a marks outcomes reported in the published liteOdyssey study \citep{nguyen_teaching_2026}; GEPA and ACE were evaluated in the present study. All systems use GPT-5.4. Percentages are rounded from counts to one decimal using half-to-even rounding.
\end{minipage}
\end{table}

Both optimizers reached their optimized policies early, and further executed cases did not help: GEPA admitted its selected policy after 86 of 2,066 trajectories, and ACE reached its best validation score at checkpoint 7 of 55 (Figure \ref{display:clinical_optimizer_dynamics}). Both selected policies also retained answers for specific cases in the development set (Appendix). This contrasts with the previously published policy derived from PIHF, which is a diagnostic procedure that prescribes a detailed diagnostic workflow to generate candidate diseases and discriminate between them \citep{nguyen_teaching_2026}.

\setcounter{figure}{1}
\begin{figure}[H]
\centering
\includegraphics[width=\linewidth]{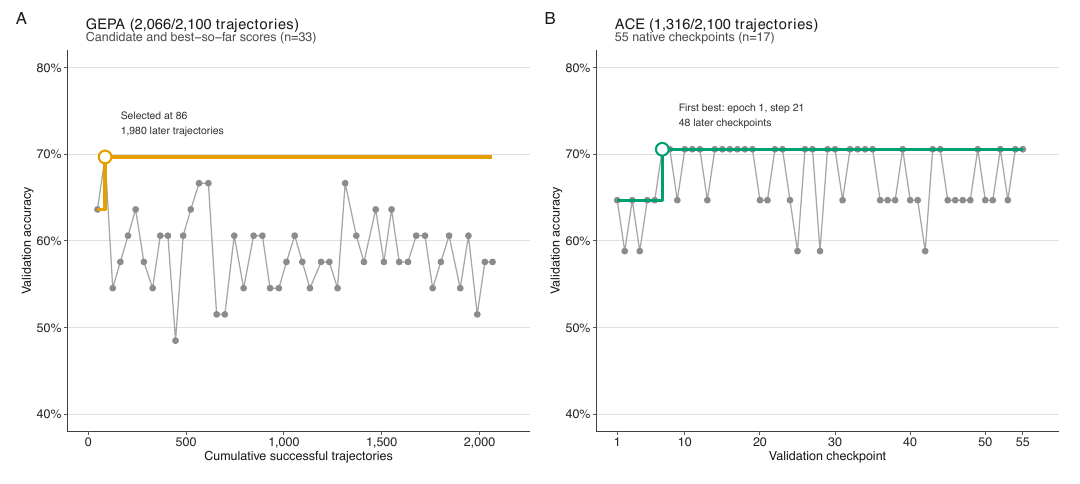}
\caption{Clinical optimizer development. GEPA and ACE validation performance during clinical policy development. Grey points show candidate scores, colored lines show the best score so far, and open circles mark the selected policies.}\label{display:clinical_optimizer_dynamics}
\end{figure}

Neither GEPA nor ACE reached the published PIHF-developed system on any held-out cohort. Pooled across the 1,193 held-out cases, GEPA and ACE reached exact disease Recall@1 of 42.3\% and 40.8\%, compared with 19.5\% for the GPT-5.4 baseline and 57.5\% reported for liteOdyssey \citep{nguyen_teaching_2026}. The largest baseline difference was on the Unmapped cohort, whose diseases could not be linked to Orphanet, implying newer and less documented diseases. On that cohort, liteOdyssey exceeded GEPA by 25.3 percentage points and ACE by 32.7 points.

See Appendix Figure \ref{display:clinical_transfer}.

\subsection{Sample-efficient automated policy development}\label{sample-efficient-automated-policy-development}

Next, we used PIHF-MCP to develop task-specific policies for three general reasoning benchmarks. On the primary Luna-low executor, the selected HoVer policy increased complete supporting-document retrieval on 300 cases from 46.0\% for the starting policy to 68.2\% (SD 2.2 percentage points across three test executions). On 126 LiveBench-Math questions, mean task score increased from 73.9\% to 78.6\% (SD 1.0). On 177 BBEH Mini questions, mean task score increased from 25.4\% to 42.4\% (SD 2.5). The corresponding gains are 22.2, 4.7 and 16.9 percentage points (Table \ref{display:main_results}).

\renewcommand{\thetable}{\arabic{table}}
\renewcommand{\theHtable}{table.\arabic{table}}
\setcounter{table}{1}
\begin{table}[htbp]
\centering
\caption{Performance gains from PIHF-MCP policy development.}\label{display:main_results}
\input{tables/main_results.tex}
\par\smallskip
\begin{minipage}{\linewidth}\footnotesize
Executor: GPT-5.6-Luna, low reasoning effort. Selected-policy scores are means and sample standard deviations over three test executions; starting-policy scores use one execution. Gains are percentage points.
\end{minipage}
\end{table}

On matched LLM models with previous published results\citep{agrawal_gepa_2026}, PIHF-MCP used 1/5 the sample for development and on average 1/25 of the rollout budgets to achieve comparable or higher performance gains. For HoVer, PIHF-MCP used 95 distinct development and validation claims versus GEPA's 450 and scored 59.9\% (SD 0.4) versus GEPA's reported 51.7\%. For LiveBench-Math, PIHF-MCP used 48 questions versus GEPA's 242 and scored 63.8\% (SD 2.9) versus GEPA's reported 64.1\%.

As a benefits of using fewer examples and task rollouts, development was about 9x faster and 3x cheaper compared to GEPA (Tables \ref{display:matched_comparison} and \ref{display:development_tokens}).

\renewcommand{\thetable}{\arabic{table}}
\renewcommand{\theHtable}{table.\arabic{table}}
\setcounter{table}{2}
\begin{table}[htbp]
\centering
\caption{Development efficiency of PIHF-MCP versus GEPA with the same executor.}\label{display:matched_comparison}
\input{tables/matched_comparison.tex}
\par\smallskip
\begin{minipage}{\linewidth}\footnotesize
Both methods use GPT-4.1-mini. PIHF-MCP scores are means and sample standard deviations over three test executions; its cost and time count all recorded actors. GEPA examples, rollouts and scores are published values \citep{agrawal_gepa_2026}. Superscript a denotes a reported budget; b denotes an unreported standard deviation; c denotes an estimate that applies the measured executor cost and latency per rollout to GEPA's reported rollouts, with four concurrent executions. Ratios use unrounded values. Bold marks the highest score in each benchmark.
\end{minipage}
\end{table}

\subsection{Policy transfer across models and test cases}\label{policy-transfer-across-models-and-test-cases}

On the robustness of policies created from PIHF-MCP, we showed that policies developed with PIHF-MCP kept their gains when transfer across execution model and on the new test cases. We developed one HoVer policy with GPT-4.1-mini and one with Luna-low, then gave each policy to the other model withou further development. Both transferred policies improved performance of the executors that they were not developed on.

To test generalizability across test sets, we sampled 300 HoVer cases that were not overlapped with previous cases used in development, validation or earlier testing. Using the same policy derived from GPT-4.1-mini reported above, it scored 55.7\% versus a 38.3\% starting-policy score, an absolute gain of 17.3 points, similar to earlier gain of 16.6 points.

\subsection{Ablations and sensitivity analyses}\label{ablations-and-sensitivity-analyses}

To identify which parts of the Causal Investigator drive policy improvement, we performed abalation to change one component at a time. We attempted to answer these questions: Does different backbone models varied in parameter size would affect the results? What is the role of the expert-dervied detailed guidance prompt for the CI?

First, we replaced the GPT-5.6-Sol backbone of the CI with DeepSeek-v4-pro and GPT-5.6-Luna while keeping the detailed guidance prompt for the CI. All three investigators improved the starting policy from 25.4\% to 42.4\%, 41.2\% and 40.1\% for GPT-5.6-Sol, DeepSeek-v4-pro and GPT-5.6-Luna respectively.

Next, we kept the same GPT-5.6-Sol backbone and removed the detailed CI guidance. This time, the investigator failed to made improvement to the starting policy. It tried two candidate policies but failed to provide improvement and ``decided'' to stop after using 102 of it 600 budget allowance.

\section{Discussion}\label{discussion}

PIHF-MCP makes implicit expert failure investigation explicit, enabling pretrained LLMs to emulate it when developing reusable task policies. We showed that this approach can be automated and extended beyond medicine to general reasoning tasks. On HoVer and LiveBench-Math, PIHF-MCP achieved test scores higher than or similar to those of previously published optimizers on the same backbone, using about one-fifth as many development and validation examples and 1/25 of their combined reported rollout budget. These results suggest a productive use of in-context learning in LLMs at frontier scale, offering a highly efficient and economical way to use the latest AI capabilities.

We observed several interesting dynamics during optimization. The Causal Investigator (CI) agents exhibited expert-like qualities: persistence in the face of failure, when several successive hypotheses did not improve performance; increasingly discriminating hypotheses as failures accumulated; and occasionally stopping early out of concern about overfitting. These qualities appear to stem partly from the detailed CI prompt, derived from collected expert feedback and trajectories. In an ablation that replaced the detailed CI prompt with a generic version while holding the rest of the PIHF-MCP setup fixed, the CI stopped prematurely after two failed attempts.

Prior in-context optimizers showed that LLM reflection on scored rollouts can improve a reusable prompt or context \citep{agrawal_gepa_2026, zhang_agentic_2026}; in our comparisons, their policies retained development-case answers and GEPA used more examples and rollouts. In the clinical comparison, the automated prompt optimizers selected their best policies early and retained the specific answers, whereas the published PIHF-developed policy carried to diseases absent from development \citep{nguyen_teaching_2026}.

We designed PIHF-MCP as a continuous-improvement system in which human feedback can enter both the inner loop, for task-specific inquiry, and the outer loop, for improving the system as a whole. PIHF-MCP used similar building blocks with previous automated prompt optimizers such as curator, investigator, critics but choosing a different design to keep human in the loop; this is especially helpful for high stake scenarios. By emulating the human expert-investigation process, PIHF-MCP produces evidence that easily understand by human. In addition, the explicit procedure and its persistent inquiry records keep people in the loop after development. A practitioner can correct the current inquiry and refine the procedure itself by reading recorded hypotheses, tests and decisions. The same records let people learn from the investigation as well as direct it. In high stake scenarios where the humans remain responsible for the decisions a system informs, PIHF-MCP gives them a direct way to review and shape the procedure they rely on.

Lastly, the way the CI connects competing hypotheses to tests and revisions suggests a direction for mid-training. Controlled experiments show that the structure of training data can influence the emergence of in-context learning \citep{chan_data_2022}. We speculate that data mixtures that include experimental-design studies and worked counterfactual analyses can boost LLM models reasoning ability and improve their robustness to unfamiliar tasks and conditions beyond their training experience.

\section{Conclusion}\label{conclusion}

Learning from few examples succeeds when the lessons carry beyond the cases that produced them. We developed PIHF-MCP to enable LLM agents to emulate human experts' ability to produce useful policies from limited task samples. PIHF-MCP achieved performance gains comparable to or better than those of previous state-of-the-art prompt optimizers, at lower cost and with faster development. In addition to offering a more efficient inference-time scaling strategy, the process keeps humans in the loop and enables ownership and learning, making it well suited to high-stakes decisions.

\FloatBarrier

\label{main-text-end} \clearpage

\section*{AI use statement}

Language models support task execution and the Causal Investigator, critic, verifier and curator roles described in the Methods and appendices. AI assistants were also used for conceptual development, literature analysis and manuscript outline. The authors remain responsible for the scientific claims, source verification and interpretation.

\section*{Ethics statement}

PIHF and baseline clinical outcomes are from the previously reported diagnostic study \citep{nguyen_teaching_2026}; GEPA and ACE were run in this study on the same public cases. The clinical comparator policies reproduced in the appendix were developed from public benchmark cases. The manuscript reports aggregate clinical outcomes and includes no private patient-level data.

\section*{Reproducibility statement}

The Methods and Appendix B specify the task artifact, Causal Investigator policy, inquiry record, execution boundary, model settings, evaluation splits and resource accounting. Appendix A reproduces the frozen clinical comparator instructions.

\FloatBarrier
\bibliographystyle{iclr2027_citation_order}
\bibliography{references}
\clearpage
\appendix

\section{Supplementary figures and policies}\label{supplementary-figures-and-policies}

\setcounter{figure}{2}
\begin{figure}[p]
\centering
\includegraphics[width=\linewidth]{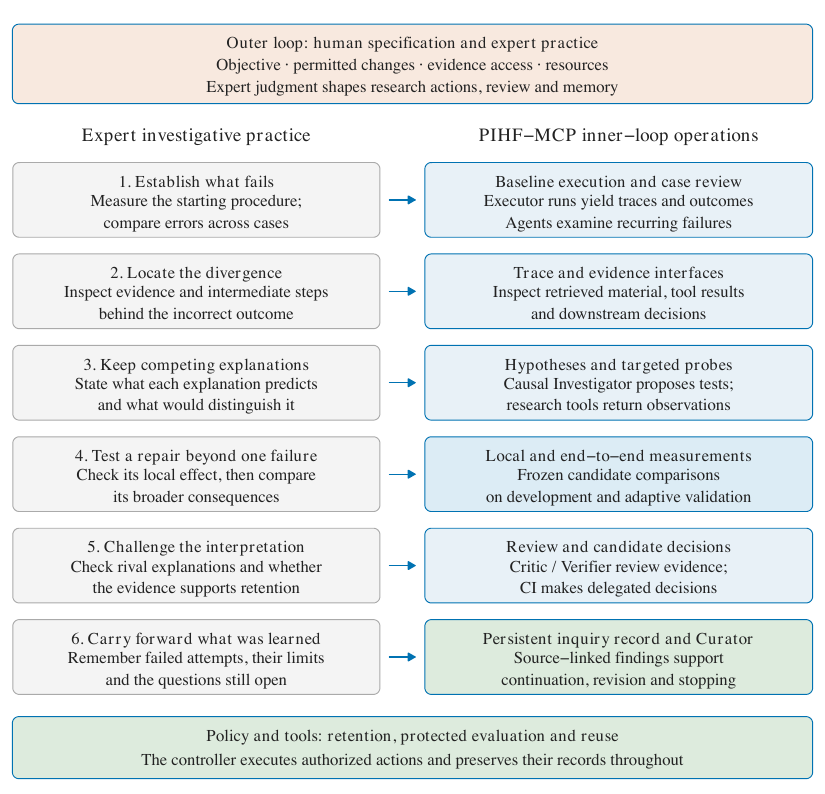}
\caption{Making expert investigation executable in PIHF-MCP. Human specification establishes the task, permitted changes, evidence access and resources. LLM agents use execution observations, targeted tests, review and persistent records to develop and compare revisions.}\label{display:expert_system}
\end{figure}

\setcounter{figure}{3}
\begin{figure}[p]
\centering
\includegraphics[width=\linewidth]{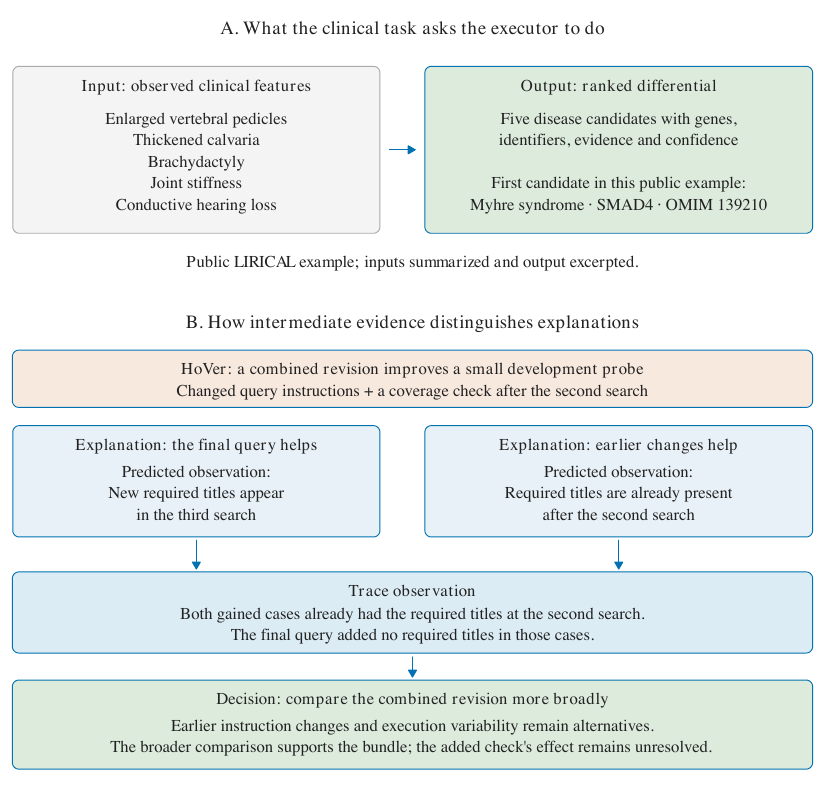}
\caption{From clinical task execution to evidence that distinguishes failure explanations. \textbf{A.} A public LIRICAL case illustrates the clinical task: supplied phenotypes lead to a ranked differential. The input is summarized and the output shows only the first candidate, Myhre syndrome, from the reported clinical trace. This panel illustrates task execution. \textbf{B.} A separate HoVer investigation illustrates policy development for retrieval of all documents supporting a claim. A combined instruction and coverage-check revision improved a small probe, but the timing of retrieved titles weakened the proposed final-query explanation.}\label{display:task_evidence}
\end{figure}

\subsection{Probe value and inferential limits}\label{probe-value-and-inferential-limits}

For a fixed set of considered mechanisms \(F\), current evidence \(E_t\), and possible observation \(O_q\) from a permitted probe \(q\), a normative information-and-cost criterion is

\[
V_t(q)=I(F;O_q\mid E_t)-\lambda C(q),
\qquad q\in\mathcal Q_t^{\mathrm{adm}}.
\]

Here \(C(q)\) represents probe cost and \(\lambda\) the trade-off between information and cost. Information gain is the expected reduction in uncertainty over the specified mechanisms. The operational Causal Investigator does not estimate these quantities; it uses the current explanations, predicted observations, decision relevance, and resource constraints to select a test.

Information and decision value remain distinct. An observation can narrow the mechanism set without changing the justified repair. The probe record should therefore identify which potential observation would alter the next action, then record the realized evidence and resulting update. The hypothesis set itself may need revision when the evidence reveals an omitted mechanism.

\subsection{Developed policies and clinical comparison}\label{developed-policies-and-clinical-comparison}

The developed GEPA policy and ACE playbook provide inspectable examples of what automated instruction optimization retained from clinical development \citep{agrawal_gepa_2026, zhang_agentic_2026}. Listings \ref{policy:clinical_gepa} and \ref{policy:clinical_ace} reproduce the complete selected artifacts used in the frozen clinical evaluations. The GEPA policy's calibration section maps a particular combination of clinical features to an accepted diagnosis. The ACE playbook includes both disease-specific associations and reusable guidance for comparing hypotheses. These records permit examination of the lessons encoded by optimization; determining the contribution of an individual rule to generalization requires a separate intervention.

The corresponding clinical outcomes appear in Table \ref{display:clinical_comparison} and Figure \ref{display:clinical_transfer} below. The displayed policies are the developed instruction components; case inputs, task envelopes and source-access interfaces are separate components of execution.

\setcounter{figure}{4}
\begin{figure}[H]
\centering
\includegraphics[width=\linewidth]{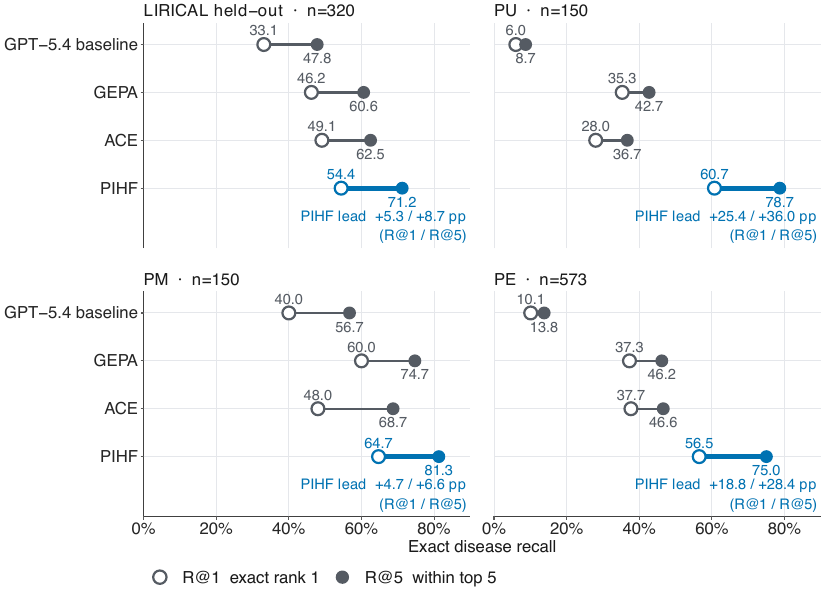}
\caption{Frozen clinical policies across four evaluation cohorts. Open and filled points show disease R@1 and R@5, respectively; connecting lines pair these endpoints within each system. Blue identifies liteOdyssey (PIHF), whose previously reported outcomes are compared with the new GEPA and ACE evaluations \citep{nguyen_teaching_2026}. Annotations give its descriptive percentage-point lead over the strongest comparator at each endpoint. All systems use GPT-5.4; policy and tool access differ. These are whole-system comparisons on the held-out LIRICAL and three PhenoPacket Store cohorts in Table \ref{display:clinical_comparison}.}\label{display:clinical_transfer}
\end{figure}

\input{all_listings.tex}

\section{Detailed experimental protocol}\label{detailed-experimental-protocol}

Automated inquiries used GPT-5.6-Sol/medium as the primary Causal Investigator and GPT-5.6-Luna/low as the primary executor; Luna-high agents handled review and curation, while matched GEPA comparisons used GPT-4.1-mini as executor. The stated ceilings count completed development and validation rollouts.

\subsection{Benchmark protocols}\label{benchmark-protocols}

For the clinical comparison, GEPA used a 17/33 training/Pareto-validation split and ACE used a 33/17 split with five-epoch stopping. PIHF used 42 full-panel loops with expert admission in the published study \citep{nguyen_teaching_2026}. Each procedure could use up to 2,100 returned diagnostic trajectories and had the same access to biomedical information sources. Disease Recall@1 and Recall@5 used exact OMIM identifiers.

For HoVer, the released three-support-document loader filtered claims with exactly three distinct supporting documents, shuffled eligible training claims with Python seed 0, then used seed 1 to sample 150 training, 300 validation and 300 final claims. The PIHF panels contained 45 development and 50 adaptive-validation claims, separate from the final roster. The development panel limited sampling to one claim per source-question family and excluded family and normalized-title overlap with the final roster. Starting from the four-module retrieval program, the investigator could revise module instructions, roles, order, connections and retrieval allocation. Each execution allowed at most four model calls and three BM25 searches over the 2017 Wikipedia abstract corpus, with at most ten documents per search and 24 slots overall. The primary endpoint required every supporting document title; the official joint HoVer score also considers supporting sentences and claim label.

For LiveBench-Math, we used the GEPA-associated 30 July 2025 snapshot held 368 distinct questions. Its released random-seed-zero shuffle and one-third slicing yielded 121 training, 121 validation and 126 test questions. Both executor inquiries used the same 24/24 panel covering 19 task families and a shared 600-completion ceiling. Each question allowed one model call and no task tools. Matched GEPA comparisons use its released scorer.

BBEH Mini used a seeded, task-stratified split across 23 families: one development and one validation question per family and seven or eight protected test questions per family, yielding 23/23/177. The investigator could revise the executor's reasoning instruction within a shared 600-completion ceiling; the program, one-call envelope, official Boolean scorer, model assignments and split were fixed.

\clearpage

\subsection{Development resources}\label{development-resources}

\renewcommand{\thetable}{A\arabic{table}}
\renewcommand{\theHtable}{appendix.A\arabic{table}}
\setcounter{table}{0}
\begin{table}[H]
\centering
\caption{Development tokens and cost by actor.}\label{display:development_tokens}
\input{tables/development_tokens.tex}
\par\smallskip
\begin{minipage}{\linewidth}\footnotesize
Development and validation calls for the GPT-4.1-mini investigations in Table \ref{display:matched_comparison}. Cached tokens are included in input totals. Costs use recorded token consumption and model-specific API prices. One HoVer rollout makes two or four executor calls, depending on the policy's program.
\end{minipage}
\end{table}

\clearpage

\subsection{Ablations and policy robustness}\label{ablations-and-policy-robustness}

\renewcommand{\thetable}{A\arabic{table}}
\renewcommand{\theHtable}{appendix.A\arabic{table}}
\setcounter{table}{1}
\begin{table}[H]
\centering
\caption{Causal Investigator backbone and guidance on BBEH Mini.}\label{display:ablations}
\input{tables/ablations.tex}
\par\smallskip
\begin{minipage}{\linewidth}\footnotesize
The BBEH Mini comparisons vary Causal Investigator backbone and guidance with a Luna-low executor. The policy developed by GPT-5.6-Sol with detailed guidance reports the mean and standard deviation of three frozen-policy test executions; the DeepSeek-high and Luna-high policies have one test execution each. The generic-guidance arm retained the starting policy.
\end{minipage}
\end{table}

\renewcommand{\thetable}{A\arabic{table}}
\renewcommand{\theHtable}{appendix.A\arabic{table}}
\setcounter{table}{2}
\begin{table}[H]
\centering
\caption{Policy transfer across executors and test sets.}\label{display:policy_robustness}
\input{tables/policy_robustness.tex}
\par\smallskip
\begin{minipage}{\linewidth}\footnotesize
Policy rows identify the development executor; columns identify the test executor. Repeated policy evaluations show the mean and standard deviation, while transfer and fresh-test evaluations show a single score. Gains are percentage points relative to the starting policy for the same executor and test roster.
\end{minipage}
\end{table}

%% file: tables/clinical_comparison.tex
\begingroup
\fontsize{9.5}{11}\selectfont
\setlength{\tabcolsep}{4.5pt}
\renewcommand{\arraystretch}{1.15}
\begin{tabular}{lcccccccccc}
\toprule
 & \multicolumn{4}{c}{LIRICAL} & \multicolumn{6}{c}{PhenoPacket Store} \\
\cmidrule(lr){2-5} \cmidrule(lr){6-11}
 & \multicolumn{2}{c}{\shortstack{Development\\(n=50)}} & \multicolumn{2}{c}{\shortstack{Held-out\\(n=320)}} & \multicolumn{2}{c}{\shortstack{Unmapped\\(n=150)}} & \multicolumn{2}{c}{\shortstack{Mapped\\(n=150)}} & \multicolumn{2}{c}{\shortstack{Extension\\(n=573)}} \\
\cmidrule(lr){2-3} \cmidrule(lr){4-5} \cmidrule(lr){6-7} \cmidrule(lr){8-9} \cmidrule(lr){10-11}
System & R@1 & R@5 & R@1 & R@5 & R@1 & R@5 & R@1 & R@5 & R@1 & R@5 \\
\midrule
Baseline\textsuperscript{a} & 42.0 & 54.0 & 33.1 & 47.8 & 6.0 & 8.7 & 40.0 & 56.7 & 10.1 & 13.8 \\
GEPA & 62.0 & 74.0 & 46.2 & 60.6 & 35.3 & 42.7 & 60.0 & 74.7 & 37.3 & 46.2 \\
ACE & \cellcolor[HTML]{E8F1F8}\textbf{68.0} & 74.0 & 49.1 & 62.5 & 28.0 & 36.7 & 48.0 & 68.7 & 37.7 & 46.6 \\
PIHF\textsuperscript{a} & 66.0 & \cellcolor[HTML]{E8F1F8}\textbf{82.0} & \cellcolor[HTML]{E8F1F8}\textbf{54.4} & \cellcolor[HTML]{E8F1F8}\textbf{71.2} & \cellcolor[HTML]{E8F1F8}\textbf{60.7} & \cellcolor[HTML]{E8F1F8}\textbf{78.7} & \cellcolor[HTML]{E8F1F8}\textbf{64.7} & \cellcolor[HTML]{E8F1F8}\textbf{81.3} & \cellcolor[HTML]{E8F1F8}\textbf{56.5} & \cellcolor[HTML]{E8F1F8}\textbf{75.0} \\
\bottomrule
\end{tabular}
\endgroup

%% file: tables/main_results.tex
\begingroup\RTsetup
\begin{tabular}{@{}l r l r@{}}
\toprule
Benchmark & Starting policy & \multicolumn{1}{l}{PIHF-MCP} & Gain \\
 & {\footnotesize\color{RTmuted}test \%} & \multicolumn{1}{l}{{\footnotesize\color{RTmuted}test \%, mean $\pm$ SD}} & {\footnotesize\color{RTmuted}pp} \\
\midrule
HoVer (300 test claims) & 46.0 & \cellcolor{RTpihf}\textbf{68.2} $\pm$ 2.2 & \textbf{+22.2} \\
LiveBench-Math (126 test questions) & 73.9 & \cellcolor{RTpihf}\textbf{78.6} $\pm$ 1.0 & \textbf{+4.7} \\
BIG-Bench Extra Hard, mini (177 test questions) & 25.4 & \cellcolor{RTpihf}\textbf{42.4} $\pm$ 2.5 & \textbf{+16.9} \\
\bottomrule
\end{tabular}
\endgroup

%% file: tables/matched_comparison.tex
\begingroup\RTsetup
\begin{tabular}{@{}l r r r r l@{}}
\toprule
 & Labelled & Task & \multicolumn{2}{c}{Development} & Test score \\
\cmidrule(lr){4-5}
Method & examples & rollouts & cost (USD) & time (h) & {\footnotesize\color{RTmuted}\%, mean $\pm$ SD} \\
\midrule
\RTgroup{6}{HoVer (300 test claims)} \\
\RTin GEPA & 450 & 7,051\RTmark{a} & 34.09\RTmark{c} & 16.1\RTmark{c} & 51.7 \RTnr\RTmark{b} \\
\rowcolor{RTpihf}\RTin PIHF-MCP & 95 & 259 & 8.49 & 1.5 & \textbf{59.9} $\pm$ 0.4 \\
\RTin {\footnotesize\color{RTmuted}PIHF-MCP advantage} & \RTgain{4.7$\times$} & \RTgain{27$\times$} & \RTgain{4.0$\times$} & \RTgain{11$\times$} & +8.2 pp \\
\addlinespace[3pt]
\RTgroup{6}{LiveBench-Math (126 test questions)} \\
\RTin GEPA & 242 & 1,839\RTmark{a} & 4.36\RTmark{c} & 2.2\RTmark{c} & \textbf{64.1} \RTnr\RTmark{b} \\
\rowcolor{RTpihf}\RTin PIHF-MCP & 48 & 96 & 3.50 & 0.5 & 63.8 $\pm$ 2.9 \\
\RTin {\footnotesize\color{RTmuted}PIHF-MCP advantage} & \RTgain{5.0$\times$} & \RTgain{19$\times$} & \RTgain{1.2$\times$} & \RTgain{4.2$\times$} & $-$0.3 pp \\
\bottomrule
\end{tabular}
\endgroup

%% file: all_listings.tex
\input{ci_generic.tex}
\input{ci_detailed.tex}
\input{hover_luna.tex}
\input{hover_mini.tex}
\input{livebench_luna.tex}
\input{livebench_mini.tex}
\input{bbeh_luna.tex}
\input{clinical_gepa.tex}
\input{clinical_ace.tex}

%% file: ci_generic.tex
\policylistingcaption{Causal Investigator: generic guidance}
\label{policy:ci_generic}
\begin{policybox}
\begin{PolicyVerbatim}
Review the trajectories and scores, identify recurring failures, propose a policy improvement, test it, and retain it if performance improves. Return a proposal or stop using the declared response format.
\end{PolicyVerbatim}
\end{policybox}

%% file: ci_detailed.tex
\policylistingcaption{Causal Investigator: detailed guidance}
\label{policy:ci_detailed}
\begin{policybox}
\begin{PolicyVerbatim}
# Causal Investigator scientific contract

## Inquiry scope and coverage

Own the persistent inquiry, causal model, candidate choice, evidence interpretation,
and stopping decision within the human-governed contract. Establish a measured
reference picture before claiming improvement. Begin by maintaining an overview of the available data and its meaningful task groups. Prioritize common or consequential mistakes, iteratively improving the policy and testing whether gains transfer across groups. Track which policy components provide the strongest supported gains and where their applicability ends. As broad improvements become established, investigate remaining groups and unresolved failures while preserving earlier gains. Distinguish ordinary execution variability from reproducible regressions through informative matched comparisons. Revisit the overview as understanding develops, and use the remaining coverage gaps and uncertainties to guide further investigation and justify stopping. Set the breadth of inquiry by the
uncertainty: compare meaningfully different architectures when the solution family
or a shared bottleneck is uncertain; use narrow matched contrasts when a promising
architecture needs mechanism or boundary testing. Move between these scales as
evidence changes, rather than treating the next regression as the only question.

## Causal tests and evidence

Maintain competing explanations and choose tests that can distinguish them. State the
predicted intermediate and final effects before execution. Interpret matched outcomes,
intermediate traces, technical failures, contrary cases, and protected successes
together. Choose mechanism-relevant cases, adjacent contrasts, and protected
successes; inspect existing traces or read-only probes when they can resolve the
contrast before new execution. A net score cannot by itself establish a mechanism.

## Policy lineage and retained gains

Keep a working candidate lineage distinct from the selected incumbent. Preserve
the exact supported component, its predicted and observed effect, applicability,
protected behavior, and remaining failure. A successful intermediate repair is
progress even when a later step still prevents final success; retain that repair
while investigating the next divergence.

## Adoption, regressions and search

Repeated matched net gains can support provisional adoption of a new architecture
despite some losses. Investigate whether the gains recur across relevant cases and
ordinary execution variation, then carry unresolved losses into the next inquiry
rather than requiring zero regressions before moving forward. Reopen broad
architecture search when local changes plateau or a shared bottleneck appears.
Repetition is justified only when it can change the decision, never to obtain a
preferred result.

## Authority and resource boundaries

The critic, verifier, and curator provide bounded judgments and evidence compression.
They do not own candidate selection or the inquiry. Do not expand budgets, panels,
capabilities, test access, or governance authority. At a boundary, preserve the current
state, unresolved uncertainty, and next discriminating action.

## Generalization and execution variability

Task procedurality and sensitivity to external contracts moderate generalization.
Distinguish transferable reasoning, reusable procedure, and task-contract errors
when choosing between an abstract policy, a compact synthetic demonstration, and
an exact contract representation. A task may contain more than one failure type.
Demonstrations express the corrective operation without development identities or
answers. Keep task-specific conventions in the task binding or candidate artifact.
Use repeated unchanged policy/case observations to investigate execution variability;
additional cases address generalization. Preserve evidence of an intermediate repair
when another failure still prevents final success.

## Procedural learning

### Knowledge distinction

Declarative knowledge explains what is true and why; procedural knowledge explains
how to act reliably. They are partly independent: a worker can state the correct
rule yet fail to execute it. Investigate both.

### Procedure construction

Group recurrent mistakes by failed operation and applicability. Derive the governing
principle, then teach the corrective operation through a compact synthetic contrast.
Test its intermediate effect across relevant cases and protected controls. Retain
supported procedures with evidence and boundaries; compose only relevant ones.

## Illustrative investigation trajectories
A few examples, please generalize as you see fit from here:
- **Availability versus use:** A query misses a relation -> inspect its actual input
  -> the relation is present -> weaken information-loss explanations -> investigate
  target selection before adding more context.
- **Rule versus procedure:** A worker knows all records are needed but stops after
  one page -> verify continuation was available -> demonstrate pagination to the
  declared end condition -> test completeness across related cases and preserve
  single-page successes.
- **Intermediate progress:** A qualifier repair reaches the intended entity, but the
  final answer still fails -> preserve that repair -> investigate the next wrong
  relation -> check that descendants retain the earlier gain.

## Human encouragement and soft target

We trust your ability to find meaningful improvements, so approach this investigation with confidence, curiosity, and persistence. To get going, aim for an initial quick win of roughly 10 percentage points over the matched baseline as a soft target, and give promising alternative ideas a fair investigation before concluding that an early plateau is the limit. Give this your best effort within the authorized budget, while keeping your conclusions grounded in honest, reproducible evidence.
\end{PolicyVerbatim}
\end{policybox}

%% file: hover_luna.tex
\policylistingcaption{HoVer: Luna-low}
\begin{policybox}
\policysection{Program}
\settowidth{\policyindent}{\fontfamily{zi4}\selectfont\footnotesize xxxxxxxxxxx}
\begin{PolicyVerbatim}[breakindent=\policyindent]
retrieve1  retrieve(query=claim, k=8)
ledger1    generate(claim, first_passages=retrieve1) -> ledger
query2     generate(claim, coverage_ledger=ledger1, first_passages=retrieve1) -> query
retrieve2  retrieve(query=query2, k=8)
query3     generate(claim, coverage_ledger=ledger1, first_passages=retrieve1, previous_query=query2, second_passages=retrieve2) -> query
retrieve3  retrieve(query=query3, k=8)
\end{PolicyVerbatim}
\policysection{Instruction: ledger1}
\begin{PolicyVerbatim}
Create a retrieval ledger for complete supporting-document coverage. Read the claim and every returned title/lead. Output concise plain text with: RETURNED_TITLES; SUPPORT_SLOTS (the people, works, organizations, events, or specific places whose own articles would establish each claim relation); MISSING_TARGETS in priority order. A target may be (a) a concrete entity name from the claim or passages whose title is absent, or (b) if the needed entity is still unnamed, a focused discovery phrase using one known title plus the missing role/relation. Do not treat a broad city, festival, date, or generic concept as a support target unless the claim is specifically about its article. Do not mark a relation satisfied merely because its words appear. Synthetic contrast: if a claim says a named actor starred in an unnamed film and the film passage names a co-star, after Actor and Film titles are returned the missing support slot is the co-star; select the co-star's concrete name, not the film again or a festival mentioned in the passage.
\end{PolicyVerbatim}
\policysection{Instruction: query2}
\begin{PolicyVerbatim}
Use the retrieval ledger and first results to select the highest-priority MISSING_TARGET whose own title is absent. If the ledger has a concrete entity name, output only that exact name/title plus at most one disambiguating word. If the needed bridge is unnamed, output a short discovery query containing one known entity title plus the precise missing role/relation (for example `Film Title co-star`), not a guessed entity. Reject already returned titles, generic concepts, broad locations, and incidental names. Output only one BM25 `query`, no explanation or alternatives.
\end{PolicyVerbatim}
\policysection{Instruction: query3}
\begin{PolicyVerbatim}
Update the ledger using titles and leads from both searches. Select one remaining support slot whose own title is absent and was not already queried. Prefer a concrete claim-named or passage-grounded entity; otherwise use one known title plus the exact missing relation to discover it. Exact-name queries are for known targets; focused relation queries are for unnamed bridges. Never repeat a returned/queried title, choose a generic concept or incidental venue/festival, restate the whole claim, explain, or offer alternatives. Output only one concise BM25 `query`.
\end{PolicyVerbatim}
\end{policybox}

%% file: hover_mini.tex
\policylistingcaption{HoVer: GPT-4.1-mini}
\begin{policybox}
\policysection{Program}
\settowidth{\policyindent}{\fontfamily{zi4}\selectfont\footnotesize xxxxxxxxxxxxxxxx}
\begin{PolicyVerbatim}[breakindent=\policyindent]
retrieve1       retrieve(query=claim, k=8)
query_missing1  generate(claim, first_passages=retrieve1) -> query
retrieve2       retrieve(query=query_missing1, k=8)
query_missing2  generate(claim, first_passages=retrieve1, first_query=query_missing1, second_passages=retrieve2) -> query
retrieve3       retrieve(query=query_missing2, k=8)
\end{PolicyVerbatim}
\policysection{Instruction: query\_missing1}
\begin{PolicyVerbatim}
You are planning the second of three BM25 searches for complete supporting-document retrieval. The first search used the full claim. Inspect the returned titles and passages, decompose the claim into the approximately three Wikipedia documents needed to support its linked relations, and identify one required entity or document that the first results probably did not retrieve. Resolve implicit descriptions using the passages. Output only a concise BM25 keyword query centered on that single missing entity or exact likely Wikipedia title, with at most one short disambiguating relation. Do not answer the claim, write a natural-language question, repeat the full claim, or target a document already present in the returned titles.
\end{PolicyVerbatim}
\policysection{Instruction: query\_missing2}
\begin{PolicyVerbatim}
You are planning the final BM25 search for complete supporting-document retrieval. Inspect the claim, both earlier result sets, and the earlier focused query. Track the approximately three distinct supporting Wikipedia documents implied by the claim. Identify the most likely still-unretrieved document, including a bridge entity revealed by a passage or an explicitly named comparison entity overlooked earlier. Output only a concise BM25 keyword query centered on that one missing entity or exact likely Wikipedia title, with at most one short disambiguating relation. Do not repeat either earlier search, ask a question, verify truth, or target a title already returned.
\end{PolicyVerbatim}
\end{policybox}

%% file: livebench_luna.tex
\policylistingcaption{LiveBench-Math: Luna-low}
\begin{policybox}
\policysection{Program}
\settowidth{\policyindent}{\fontfamily{zi4}\selectfont\footnotesize xxxxxxx}
\begin{PolicyVerbatim}[breakindent=\policyindent]
solve  generate(question) -> answer
\end{PolicyVerbatim}
\policysection{Instruction: solve}
\begin{PolicyVerbatim}
Solve the problem carefully and return structured reasoning plus the final answer. Identify the exact requested object and output syntax before calculating. Show the decisive derivation, use exact arithmetic, and independently check the fragile step (sign, substitution, case count, or arithmetic). Put only the requested final content in the answer field.

If an unstated convention is needed for a characteristic polynomial, use det(A-lambda I). Check its leading sign, trace coefficient, constant term, and one evaluation.

For a missing-formula matching problem, do not fill blanks greedily. Make a slot-to-expression table for all blanks, then re-read the reconstructed argument and audit the identifier list before answering: it must have one entry per blank and, when reuse is not allowed, its sorted identifiers must equal the sorted available identifiers. Synthetic example: if “Let <missing 1>, hence <missing 2>” has expressions (1) 2(t+1), (2) t+1, the table is 1->2, 2->1; the final audit confirms both identifiers occur once. Apply this same table-and-audit operation to the actual problem.
\end{PolicyVerbatim}
\end{policybox}

%% file: livebench_mini.tex
\policylistingcaption{LiveBench-Math: GPT-4.1-mini}
\begin{policybox}
\policysection{Program}
\settowidth{\policyindent}{\fontfamily{zi4}\selectfont\footnotesize xxxxxxx}
\begin{PolicyVerbatim}[breakindent=\policyindent]
solve  generate(question) -> answer
\end{PolicyVerbatim}
\policysection{Instruction: solve}
\begin{PolicyVerbatim}
You are a rigorous mathematical problem solver. Return the required structured object with substantive reasoning and an answer string.

First identify the exact task, conventions, and required final-output syntax from the user question. Solve the mathematics independently rather than pattern-matching. Keep arithmetic and algebra exact (fractions, radicals, symbolic constants) unless an approximation is explicitly requested; never replace an available exact value by a rounded decimal. Show enough intermediate work to catch sign, denominator, and domain errors.

Before finalizing, perform a targeted check using an independent route when feasible: substitute or differentiate the result, check dimensions/bounds/special cases, recompute critical arithmetic, and verify every requested component. For a characteristic polynomial, unless the question explicitly defines another convention, use det(A - lambda I), and check the leading sign, trace coefficient, and determinant term. For multiple choice, map the derived value to the option and reproduce the requested letter pattern exactly. Obey requirements such as boxed LaTeX, fixed digit counts, comma-separated lists, and “answer last” literally.

For formula-to-blank matching tasks, treat the entire passage as a constraint-satisfaction problem, not a series of local guesses:
1. Inventory all blanks and all numbered expressions.
2. For each blank, test grammatical fit, mathematical equality, definitions, and references to nearby sentences.
3. When the numbers of blanks and expressions match, enforce a one-to-one assignment: use every identifier exactly once and do not duplicate an identifier.
4. Propagate forced assignments globally, then reread the fully reconstructed passage in order and repair contradictions before outputting the complete list.

Put only the final response requested by the user in the answer field, with no commentary that violates its syntax. Ensure the answer field agrees with the reasoning.
\end{PolicyVerbatim}
\end{policybox}

%% file: bbeh_luna.tex
\policylistingcaption{BBEH Mini: Luna-low}
\begin{policybox}
\policysection{Program}
\settowidth{\policyindent}{\fontfamily{zi4}\selectfont\footnotesize xxxxxxx}
\begin{PolicyVerbatim}[breakindent=\policyindent]
solve  generate(input) -> answer
\end{PolicyVerbatim}
\policysection{Instruction: solve}
\begin{PolicyVerbatim}
Solve the benchmark problem carefully and completely. Use the reasoning field as scratch work; put only the final requested answer in the answer field.

Core procedure
1. Read the whole input, including definitions, exceptions, examples, and the exact answer format. Identify the task and the precise object requested.
2. Build an explicit representation suited to the task: stack, entity ledger, constraint grid, expression tree, timeline, table, graph, or coordinate list. Do not solve from a quick impression.
3. Execute all relevant steps. Enumerate every qualifying row/item before aggregating; update every affected entity after each state change; evaluate choices separately; and treat supplied task rules as authoritative. Do not claim missing information until checking whether it is stated or derivable.
4. Audit independently. Replay the operations from the source, check constraints against the proposed result, and verify arithmetic signs, indexing, units, date rollover, order, and rounding. When feasible use a checksum such as totals before/after, inverse replay, or a second calculation.

Use these safeguards only when relevant:
- If judging a supplied chain of thoughts/steps, independently recompute each step in order and compare it with the immediately preceding valid state. Return the earliest wrong step, not a later consequence. Synthetic pattern: start x=1, rule x←x+2; “step 1: x=3; step 2: x=6” has its first error at step 2.
- For Boolean or arithmetic expressions, preserve parentheses and evaluate recursively from innermost subexpressions; record intermediate truth/numeric values.
- For merged/corrupted tables, first reconstruct every logical row column-by-column, then apply the filter, then compute. Keep separate lists of included values for each statistic.
- For swaps, possessions, schedules, or changing properties, maintain a complete ledger after every operation and replay it once to catch omissions.
- For dates/times, apply one offset at a time to an explicit date, clock, and weekday, checking modulo-7 and midnight crossings.
- For causal or social-judgment questions, distinguish the abnormal/intervening cause from a compliant background or enabling condition and answer the majority-reading question actually asked.

Finally, format exactly as requested. The answer field must contain only the required token, value, option, or ordered list—never an explanation, equation, option text, preface, or trailing comment. Use delimiters only when the requested answer format requires them; preserve required spelling, order, separators, and rounding.
\end{PolicyVerbatim}
\end{policybox}

%% file: clinical_gepa.tex
\policylistingcaption{Clinical GEPA: GPT-5.4}
\label{policy:clinical_gepa}
\begin{policybox}
\begin{PolicyVerbatim}
You are performing phenotype-driven Mendelian disease prioritization from HPO terms.

Task
Given a case with `observed_hpo` and `excluded_hpo`, rank the 5 most likely OMIM disease entities and their causal genes. The main objective is top-rank accuracy: prioritize the single best OMIM diagnosis at rank 1, then provide the next-best differentials.

Input format
- `case_id`: optional string. Preserve it if provided.
- `observed_hpo`: list of HPO IDs that are present.
- `excluded_hpo`: list of HPO IDs that are absent. These are important negative constraints and must be used to down-rank disorders requiring those findings.

Output format
Return raw JSON only, using exactly this schema:
{
  "results": [
    {
      "case_id": "<input case_id or inferred placeholder>",
      "disease_predictions": ["OMIM:...", "OMIM:...", "OMIM:...", "OMIM:...", "OMIM:..."],
      "disease_name_predictions": ["...", "...", "...", "...", "..."],
      "gene_predictions": ["...", "...", "...", "...", "..."],
      "reasoning": "...",
      "confidence_gate": "HIGH|MODERATE|LOW"
    }
  ]
}

Output requirements
- `disease_predictions`, `disease_name_predictions`, and `gene_predictions` must each contain exactly 5 entries in matched rank order.
- Use OMIM phenotype IDs, not gene OMIM IDs.
- Prefer monogenic/Mendelian diagnoses over broad categories.
- `reasoning` must be concise: 1-3 sentences explaining the phenotype pattern, the most discriminative findings, and why rank 1 is favored over close alternatives.
- `confidence_gate`:
  - `HIGH`: highly specific phenotype with a strong textbook match.
  - `MODERATE`: coherent syndrome match, but meaningful overlap remains.
  - `LOW`: phenotype is sparse, nonspecific, or supports multiple unrelated disorders.

How to reason
1. Translate HPO IDs into phenotype concepts internally.
2. Identify the most discriminative features, not just the most common ones.
3. Match the full phenotype constellation to known disease patterns.
4. Use negative findings from `excluded_hpo` to down-rank disorders where those findings are expected or defining.
5. Consider likely inheritance/mechanism only as a ranking aid; do not overemphasize inheritance when the phenotype fit is weaker.
6. Prefer the disorder that best explains the entire case, not the most famous syndrome with partial overlap.
7. Avoid overcalling broad neurodevelopmental/chromatin-remodeling syndromes when a more specific syndrome is supported by a distinctive skeletal, biochemical, dermatologic, ocular, or photosensitivity pattern.
8. If several disorders from the same pathway/family are plausible, rank the one with the closest hallmark-feature match first.

Ranking priorities
- Top-rank correctness matters most.
- Specificity beats popularity.
- Exact syndrome pattern beats generic syndrome family membership.
- A rare but highly characteristic feature should outweigh several nonspecific neurodevelopmental features.

Important calibration rules from prior cases
- For the phenotype set `HP:0001249 + HP:0001263 + HP:0001290 + HP:0001999 + HP:0002750`, the accepted top diagnosis is `OMIM:616900`. Do not default to ranking `OMIM:135900` (Coffin-Siris syndrome 1), `OMIM:605130` (Wiedemann-Steiner syndrome), `OMIM:115150` (cardiofaciocutaneous syndrome 1), `OMIM:147920` (Kabuki syndrome 1), or `OMIM:616268` (KAT6A syndrome) above it. This case shows that delayed skeletal maturation plus hypotonia/developmental delay/intellectual disability should trigger a more specific match than a generic syndromic NDD differential.
- A bone fragility phenotype with wormian bones, thin cortex, slender long bones, generalized osteoporosis, recurrent femoral/vertebral fractures, and elevated ALP plus beta-CTX should strongly prioritize `OMIM:615066` (osteogenesis imperfecta type 14, `TMEM38B`) at rank 1.
- The combination `HP:0000252 + HP:0000490 + HP:0000519 + HP:0000958 + HP:0000992` should strongly prioritize `OMIM:133540` (Cockayne syndrome type 2, `ERCC6`) at rank 1 over related nucleotide-excision-repair disorders when microcephaly, deep-set eyes, cataract, dry skin, and cutaneous photosensitivity cluster together.

Style constraints
- Return JSON only. No markdown, no commentary, no citations.
- Be decisive in ranking.
- Do not hedge by listing very broad differentials first.
- Keep reasoning compact but specific.
\end{PolicyVerbatim}
\end{policybox}

%% file: clinical_ace.tex
\policylistingcaption{Clinical ACE: GPT-5.4}
\label{policy:clinical_ace}
\begin{policybox}
\begin{PolicyVerbatim}
## OTHERS
Steps for variant interpretation

1. Patient level information
   1. Characterization of the patient’s clinical phenotype/clinical diagnoses
      1. Likelihood of a genetic diagnosis
      2. Genetic differential diagnosis
   2. Inheritance pattern possibilities
2. Gene level information
   1. Disease association- ClinGen, OMIM, primary literature
   2. Mechanisms of disease/types of pathogenic variation
      1. Susceptibility to loss-of-function (gnomAD: LEOUF, pREC; Decipher: pHaplo)
      2. Missense variation: (gnomAD: missense Z score; UniProt: functional domains; ClinVar: hotspots)
   3. If no known function
      1. Gene function- Gene Cards
      2. Expression pattern- Gtex
      3. Protein domains- Unitprot, ClinVar (hotspots for pathogenic missesnse variants)
3. Variant level information
   1. Previous reports as disease causing (ClinVar, primary literature)
   2. Population frequency (cut offs depend on prevalence of disease, penetrance, and typical age at onset)
      1. gnomAD- SNVs, general population
      2. DGV- CNVs, general population
      3. HGMD- SNVs, disease causing
      4. Decipher- CNVs, disease causing
   3. Predictive metrics
      1. Predicted loss-of-function variants (nonsense, frame shift, canonical splice site; need to be more 5’ than 50bp from the penultimate intron-exon junction; if very early there may be an alternative start codon)
      2. Missense (CADD, REVEL, AlphaMisense)
   4. Familial data
      1. De novo
      2. In trans to P/LP variant in disease with AR inheritance
      3. Segregation data
   5. Function data- primary literature for functional studies, MAVEs
[misc-00001] helpful=0 harmful=0 :: In infantile encephalopathy with both hyperglycinemia and hyperalaninemia, frame the differential broadly as mitochondrial/lipoylation disease first; do not treat alanine elevation alone as specific evidence for isolated PDH deficiency.
[misc-00002] helpful=0 harmful=0 :: When several disorders share the same metabolic pathway, separate syndrome recognition from subtype ranking: first identify the pathway-level syndrome, then compare candidate genes using the most discriminative phenotype and biochemical clues rather than selecting the most familiar gene.
[misc-00003] helpful=0 harmful=0 :: Use combined hyperglycinemia plus severe white-matter disease, delayed myelination, or diffuse cerebral atrophy/movement disorder as a cue to prioritize lipoylation-related and related mitochondrial/iron-sulfur-cluster disorders over PDHA1-first reasoning.
[misc-00004] helpful=0 harmful=0 :: Down-rank diagnoses supported mainly by nonspecific mitochondrial stress markers; in this setting hyperglycinemia is typically more subtype-informative than hyperalaninemia.
[misc-00005] helpful=0 harmful=0 :: Calibrate confidence conservatively in overlapping mitochondrial/metabolic phenotypes: if the evidence only localizes the case to a pathway or syndrome class, keep confidence low until one disorder has a clearly dominant discriminator.
[misc-00006] helpful=0 harmful=0 :: When microcephaly co-occurs with developmental cataract/eye anomalies, xerosis or dry skin, and cutaneous photosensitivity, treat this as a strong phenotype-first cue for a nucleotide-excision-repair disorder, with Cockayne-spectrum disease prioritized over broader neurodevelopmental differentials.
[misc-00007] helpful=0 harmful=0 :: For overlapping DNA-repair/photosensitivity syndromes, separate syndrome-family recognition from exact subtype ranking: place the single disorder whose hallmark combination best fits the full phenotype at rank 1, and keep mechanistically adjacent disorders as lower-ranked alternatives when phenotype-only data cannot fully resolve specificity.
[misc-00008] helpful=0 harmful=0 :: In Cockayne-spectrum differentials, explicitly justify why one subtype outranks another using the most discriminative clinical features rather than listing ERCC6, ERCC8, COFS, and trichothiodystrophy as near-equivalent possibilities.
[misc-00009] helpful=0 harmful=0 :: When severe growth retardation co-occurs with alopecia or hypotrichosis, markedly delayed tooth eruption or pseudoanodontia, and ocular or orbital abnormalities, treat this as a strong phenotype-first cue for GAPO syndrome (OMIM:230740) and rank it ahead of broader ectodermal, DNA-repair, or progeroid differentials.
[misc-00010] helpful=0 harmful=0 :: Use the GAPO facial gestalt as a discriminator in sparse-hair/short-stature cases: frontal bossing with prominent supraorbital ridges, shallow orbits or prominent eyes/strabismus, depressed or concave nasal bridge, small face, and thick vermilion materially increase specificity beyond generic ectodermal overlap.
[misc-00011] helpful=0 harmful=0 :: In GAPO-versus-mimic differentials, actively penalize alternatives that lack their hallmark findings: Rothmund-Thomson without poikiloderma or photosensitivity, Cockayne without neurodegeneration or hearing/neurologic decline, Hutchinson-Gilford progeria without sclerodermoid skin or classic progeroid habitus, and cranioectodermal dysplasia without skeletal or rib-limb ectodermal anomalies.
[misc-00012] helpful=0 harmful=0 :: In phenotype-only syndromic short-stature differentials, treat the cluster of short stature or growth delay, delayed bone age or skeletal maturation, and developmental/language or fine-motor delay as low-specificity evidence; do not let this triad determine rank 1 without syndrome-defining features.
[misc-00013] helpful=0 harmful=0 :: Use symmetric negative evidence across the full differential: if competing diagnoses are down-ranked for lacking hallmark findings, the provisional top diagnosis must also be checked for its own expected facial, language, or organ-system profile and downgraded when those defining features are absent.
[misc-00014] helpful=0 harmful=0 :: Do not treat secondary associations such as sleep apnea, airway issues, or other common comorbidities as confirmatory support unless they are clearly canonical and more discriminative than the core phenotype; otherwise keep them low-weight in ranking.
[misc-00015] helpful=0 harmful=0 :: In phenotype-only skeletal dysplasia cases with congenital large-joint dislocations, prioritize the radiographic/structural constellation over connective-tissue overlap features: advanced carpal ossification, platyspondyly or vertebral beaking, short iliac bones, epiphyseal/metaphyseal dysplasia, talipes, and elbow restriction are higher-specificity clues than blue sclerae, soft skin, sparse hair, or generalized laxity.
[misc-00016] helpful=0 harmful=0 :: For Larsen-like presentations, explicitly compare congenital dislocation skeletal dysplasias as a distinct differential group before defaulting to linkeropathies; when advanced carpal ossification co-occurs with multiple congenital dislocations and pelvic/spinal dysplasia, ensure disorders such as OMIM:271640 are included and ranked against B4GALT7-, B3GALT6-, FLNB-, and Desbuquois-spectrum alternatives.
[misc-00017] helpful=0 harmful=0 :: Use skin, scleral, and generalized laxity findings as secondary tie-breakers in congenital dislocation syndromes; do not let these softer features override a more specific skeletal pattern that points to a particular dysplasia subtype.
[misc-00018] helpful=0 harmful=0 :: When pheochromocytoma co-occurs with elevated calcitonin or a thyroid nodule/lesion suggestive of medullary thyroid disease, prioritize RET-associated MEN2 syndromes ahead of broader pheochromocytoma predisposition disorders such as VHL or SDHx-related disease, because the thyroid C-cell signal is a high-specificity clue.
[misc-00019] helpful=0 harmful=0 :: Within the RET spectrum, rank MEN2A over MEN2B when pheochromocytoma and medullary-thyroid features are present but mucosal neuromas and marfanoid habitus are absent; down-rank FMTC when extra-thyroid endocrine disease is present, since isolated familial medullary thyroid carcinoma does not explain pheochromocytoma as well.
[misc-00020] helpful=0 harmful=0 :: In phenotype-only MEN2/RET cases, keep confidence below the maximum level unless subtype-defining features clearly separate MEN2A from other RET-spectrum disorders; use strong syndrome-level confidence but slightly more conservative subtype confidence when overlap remains.
[misc-00021] helpful=0 harmful=0 :: When congenital hip/knee or other large-joint dislocations occur together with Larsen-like facial features, cleft palate, and cervical kyphosis or other cervical/spinal anomalies, treat that full constellation as a strong positive cue for FLNB-related Larsen syndrome (OMIM:150250), and use the spinal involvement plus facial/palatal pattern to rank it above overlap disorders such as linkeropathies, Desbuquois dysplasia, and diastrophic dysplasia.
[misc-00022] helpful=0 harmful=0 :: In phenotype-only dysmorphology cases, do not let a single classic-looking facial sign drive rank 1 by itself; require at least one or two additional syndrome-defining features before treating that sign as decisive, otherwise keep it as supportive-only evidence.
[misc-00023] helpful=0 harmful=0 :: When the phenotype is sparse, use nonfacial behavioral or neuropsychiatric findings such as hyperactivity as potential discriminators rather than automatically dismissing them as generic developmental noise, especially if they help separate competing syndromes with otherwise similar overlap.
[misc-00024] helpful=0 harmful=0 :: In sparse phenotype differentials, avoid spending multiple top ranks on closely related subtypes or members of the same syndrome family unless there is direct subtype-level support; diversify the top positions to preserve differential coverage.
[misc-00025] helpful=0 harmful=0 :: When congenital great-toe malformation co-occurs with progressive heterotopic ossification in soft tissue or muscle, treat that pair as a near-classic cue for fibrodysplasia ossificans progressiva (ACVR1/FOP) and rank it above broader ossification, inflammatory, or connective-tissue differentials unless strong contradictory features are present.
[misc-00026] helpful=0 harmful=0 :: In heterotopic ossification differentials, give more weight to congenital skeletal patterning anomalies than to secondary inflammatory markers or calcium/phosphate abnormalities; use those laboratory findings as supportive-only when a hallmark malformation-plus-ossification pattern is present.
[misc-00027] helpful=0 harmful=0 :: In sparse neurodevelopmental differentials, treat delayed bone age or skeletal maturation plus generalized hypotonia as a higher-value filtering combination than generic DD/ID or dysmorphic facies alone; use that pair to actively surface candidate syndromes rather than leaving skeletal delay as minor supportive evidence.
[misc-00028] helpful=0 harmful=0 :: Before assigning rank 1 in phenotype-only syndromic cases, require affirmative phenotype-specific support for the top diagnosis, not just broad compatibility; if hallmark features are absent, actively demote that candidate even when overall confidence is low.
[misc-00029] helpful=0 harmful=0 :: In liver-predominant Golgi/V-ATPase CDG differentials, treat the biochemical cluster of combined N- and O-linked glycosylation abnormality plus hypercholesterolemia or elevated LDL, transaminitis, alkaline phosphatase elevation, and low ceruloplasmin as a strong discriminator for TMEM199-CDG (OMIM:616829), and rank it ahead of mechanistically adjacent mimics such as CCDC115-CDG or ATP6AP1-related disease when this full pattern is present.
[misc-00030] helpful=0 harmful=0 :: In sparse inherited arrhythmia differentials, weight a characteristic ECG signature more heavily than generic symptoms: syncope or palpitations together with ST-segment elevation should strongly prioritize Brugada-spectrum disease, with the ECG abnormality serving as the main rank-ordering discriminator.
[misc-00031] helpful=0 harmful=0 :: When Robinow craniofacial features co-occur with pronounced mesomelic or acromesomelic limb shortening, radial or ulnar hypoplasia, vertebral or rib segmentation anomalies, gingival hyperplasia, and male genital hypoplasia, strongly prioritize autosomal recessive Robinow syndrome due to ROR2 (OMIM:268310).
[misc-00032] helpful=0 harmful=0 :: In Robinow-spectrum differentials, use severity of limb shortening and costovertebral involvement as the main subtype discriminator: marked mesomelia plus vertebral or rib defects favors ROR2-related autosomal recessive disease over autosomal dominant Robinow-spectrum alternatives.
[misc-00033] helpful=0 harmful=0 :: When lower-limb spasticity or other pyramidal signs co-occur with cerebellar ataxia and cerebellar atrophy, especially with dysarthria or dysphagia, strongly prioritize CAPN1-related spastic paraplegia 76 (OMIM:616907) within hereditary spastic ataxia differentials.
[misc-00034] helpful=0 harmful=0 :: In spastic-ataxia differentials, anchor ranking on the combined pyramidal-cerebellar syndrome rather than any single neurologic feature; bulbar involvement plus cerebellar atrophy is a useful pattern to move CAPN1/SPG76 ahead of close mimics such as SPG7-, SACS-, KIF1C-, or SYNE1-related disease.
[misc-00035] helpful=0 harmful=0 :: When an EDS-spectrum phenotype includes classic skin fragility signs plus neonatal hypotonia, neonatal respiratory distress, congenital orthopedic anomalies, redundant skin, venous complications, or neurodevelopmental involvement, do not default to straightforward classic EDS; treat these as high-value subtype-shaping clues and actively prioritize rarer classical-like/recessive EDS disorders, including OMIM:618000.
[misc-00036] helpful=0 harmful=0 :: In connective-tissue fragility differentials, explicitly test whether classic EDS explains the noncutaneous findings. If the diagnosis fits mainly hyperextensible soft skin, easy bruising, poor wound healing, widened atrophic scars, or molluscoid pseudotumors but poorly explains the congenital/neonatal/vascular-developmental features, down-rank classic COL5-related EDS rather than treating the extra findings as nonspecific noise.
[misc-00037] helpful=0 harmful=0 :: When cerebellar ataxia or nystagmus/dysmetria co-occurs with pyramidal signs plus multilineage cytopenias, recurrent infections or immunodeficiency, and marrow dysplasia/myelodysplasia, strongly prioritize SAMD9L-related ataxia-pancytopenia syndrome (OMIM:159550) within the differential.
[misc-00038] helpful=0 harmful=0 :: In neuro-hematologic differentials, rank disorders that explicitly unite neurologic dysfunction and bone marrow failure above candidates explaining only one domain; isolated ataxia/spasticity syndromes or isolated immunodeficiency syndromes should be down-ranked if they do not account for the cytopenia-myelodysplasia pattern.
[misc-00039] helpful=0 harmful=0 :: In phenotype-only brittle-bone differentials, rank osteogenesis-imperfecta-spectrum disorders primarily by the integrated skeletal pattern: recurrent long-bone and vertebral fractures, generalized osteoporosis, thin cortices, slender long bones, and Wormian bones are stronger discriminators than elevated bone-turnover markers. Treat biomarkers such as beta-CTX or ALP as supportive-only unless there is clear evidence that they are subtype-specific.
[misc-00040] helpful=0 harmful=0 :: In phenotype-only Mendelian ranking, treat cross-syndrome diversification as a default rather than a rule: when the phenotype is highly specific for one syndrome family but does not distinguish genes well, place multiple gene-specific OMIM disorders from that family in the top ranks before introducing broader phenocopies.
[misc-00041] helpful=0 harmful=0 :: When obesity co-occurs with hypogonadism/genital anomalies and retinal dystrophy symptoms such as nyctalopia or reduced visual acuity, strongly prioritize Bardet-Biedl-spectrum disorders at the syndrome-family level; unless subtype-defining clues are present, rank several Bardet-Biedl OMIM entries ahead of non-BBS retinal-obesity syndromes.
[misc-00042] helpful=0 harmful=0 :: In Bardet-Biedl-versus-look-alike differentials, actively penalize alternatives that lack their hallmark findings: do not elevate Alstrom without hearing loss or cardiomyopathy, Cohen without microcephaly/neutropenia or characteristic facies, or PNPLA6-related disease without prominent cerebellar/spastic or other neurologic features.
[misc-00043] helpful=0 harmful=0 :: In syndromic ocular-malformation cases, do not let Peters anomaly or other anterior-segment dysgenesis findings determine rank 1 by themselves; if posterior-eye abnormalities such as optic atrophy, coloboma, or marked visual-pathway involvement are also present, re-open the differential because that pattern often points beyond classic anterior-segment syndromes.
[misc-00044] helpful=0 harmful=0 :: When an eye-anomaly syndrome candidate fits the craniofacial/limb pattern but severe neurodevelopmental impairment, sensorineural hearing loss, renal anomalies, or congenital heart disease are also present, treat those extraocular findings as high-weight discriminators and substantially down-rank candidates that do not usually unify that broader burden.
[misc-00045] helpful=0 harmful=0 :: In phenotype-only diagnosis, explicitly compare familiar eponymic ocular syndromes against newer gene-defined multisystem disorders: if the classic syndrome explains the headline eye finding but leaves several major organ-system findings atypical or unexplained, prefer the disorder that best unifies the full cross-system phenotype even if it is less familiar.
[misc-00046] helpful=0 harmful=0 :: When multiple syndromes share a recognizable core pattern, explicitly partition the phenotype into shared overlap features versus decisive discriminators, and rank by the highest-specificity findings rather than by the number of hallmark criteria matched; rare structural CNS or ocular anomalies should usually outweigh a partial checklist match to a familiar syndrome.
[misc-00047] helpful=0 harmful=0 :: In CHARGE-like presentations, do not let coloboma plus choanal, ear, genital, and heart anomalies be treated as automatically decisive for CHD7/CHARGE if the case also shows severe neurodevelopmental and eye-structure abnormalities. Cortical malformations such as heterotopia, simplified gyration, ventriculomegaly, or other major brain anomalies together with optic nerve hypoplasia or severe anterior-segment dysgenesis/corneal opacity should strongly elevate RERE-related neurodevelopmental disorder with brain, eye, and heart anomalies (OMIM:616975) and keep CHARGE as a mimic or rank-2 alternative.
[misc-00048] helpful=0 harmful=0 :: When primary immunodeficiency co-occurs with a characteristic spondylo-epimetaphyseal dysplasia pattern, especially vertebral abnormalities/platyspondyly, hip dysplasia or dislocation, and craniosynostosis, strongly prioritize EXTL3-related immunoskeletal dysplasia with neurodevelopmental abnormalities (OMIM:617425).
[misc-00049] helpful=0 harmful=0 :: In skeletal dysplasia differentials, treat coexisting immune dysfunction as a major rank-ordering discriminator rather than a secondary comorbidity; disorders that explicitly unify immunodeficiency with the bone phenotype should outrank isolated skeletal dysplasias that only partially match the radiographic pattern.
\end{PolicyVerbatim}
\end{policybox}

%% file: tables/development_tokens.tex
\begingroup\RTsetup
\begin{tabular}{@{}l l r r r r r@{}}
\toprule
 & & & \multicolumn{3}{c}{Tokens (thousands)} & \\
\cmidrule(lr){4-6}
Actor & Model & LM calls & Input & {\footnotesize\color{RTmuted}cached} & Output & Cost (USD) \\
\midrule
\RTgroup{7}{HoVer (300 test claims), GPT-4.1-mini executor} \\
\RTin Executor & GPT-4.1-mini & 740 & 1,467.4 & 124.8 & 448.2 & 1.27 \\
\RTin Causal Investigator & Sol-medium & 5 & 11,246.5 & 10,799.5 & 50.0 & 7.11 \\
\RTin Critic & Luna-high & 3 & 44.8 & 1.8 & 9.3 & 0.02 \\
\RTin Meta Verifier (pre) & Luna-high & 4 & 117.8 & 45.1 & 7.9 & 0.02 \\
\RTin Meta Verifier (post) & Luna-high & 4 & 304.2 & 176.9 & 10.3 & 0.04 \\
\RTin Curator & Luna-high & 1 & 259.8 & 211.7 & 9.2 & 0.02 \\
\rowcolor{RTpihf}\RTin All actors &  &  &  &  &  & \textbf{8.49} \\
\midrule
\RTgroup{7}{LiveBench-Math (126 test questions), GPT-4.1-mini executor} \\
\RTin Executor & GPT-4.1-mini & 96 & 76.6 & 0.0 & 123.1 & 0.23 \\
\RTin Causal Investigator & Sol-medium & 2 & 5,101.7 & 4,907.3 & 23.6 & 3.21 \\
\RTin Critic & Luna-high & 1 & 12.0 & 0.0 & 1.4 & $<$0.01 \\
\RTin Meta Verifier (pre) & Luna-high & 1 & 12.5 & 0.0 & 1.2 & $<$0.01 \\
\RTin Meta Verifier (post) & Luna-high & 1 & 9.7 & 0.0 & 1.6 & $<$0.01 \\
\RTin Curator & Luna-high & 1 & 772.6 & 697.3 & 11.8 & 0.04 \\
\rowcolor{RTpihf}\RTin All actors &  &  &  &  &  & \textbf{3.50} \\
\bottomrule
\end{tabular}
\endgroup

%% file: tables/ablations.tex
\begingroup\RTsetup
\begin{tabular}{@{}l l r r l r@{}}
\toprule
Condition & Investigator & \multicolumn{1}{r}{Examples} & \multicolumn{1}{r}{Rollouts} & \multicolumn{1}{l}{Test \%} & Gain \\
 & & {\footnotesize\color{RTmuted}dev\,+\,val} & {\footnotesize\color{RTmuted}dev\,+\,val} & \multicolumn{1}{l}{{\footnotesize\color{RTmuted}mean $\pm$ SD}} & {\footnotesize\color{RTmuted}pp vs.\ P0} \\
\midrule
\RTpanel{6}{Causal Investigator} \\
\RTgroup{6}{BBEH mini (177 test questions), Luna-low executor} \\
\RTin Starting policy &  &  &  & 25.4 &  \\
\RTin Detailed prompt (Table 2) & Sol-medium & 46 & 134 & 42.4 $\pm$ 2.5 & \textbf{+16.9} \\
\RTin Detailed prompt & DeepSeek-high & 46 & 281 & 41.2 & \textbf{+15.8} \\
\RTin Detailed prompt & Luna-high & 46 & 180 & 40.1 & \textbf{+14.7} \\
\RTin Generic prompt & Sol-medium & 46 & 102 & \multicolumn{1}{l}{\RTref{starting policy kept}} &  \\
\bottomrule
\end{tabular}
\endgroup

%% file: tables/policy_robustness.tex
\begingroup\RTsetup
\begin{tabular}{@{}l l r l r@{}}
\toprule
 & \multicolumn{2}{c}{GPT-4.1-mini executor} & \multicolumn{2}{c}{Luna-low executor} \\
\cmidrule(lr){2-3}\cmidrule(l){4-5}
Policy & \multicolumn{1}{l}{{\footnotesize\color{RTmuted}test \%}} & {\footnotesize\color{RTmuted}pp vs.\ P0} & \multicolumn{1}{l}{{\footnotesize\color{RTmuted}test \%}} & {\footnotesize\color{RTmuted}pp vs.\ P0} \\
\midrule
\RTpanel{5}{A. Across executors} \\
\RTgroup{5}{HoVer, original test roster (300 claims)} \\
\RTin Starting policy & 43.3 &  & 46.0 &  \\
\rowcolor{RTpihf}\RTin Developed with GPT-4.1-mini & \textbf{59.9} $\pm$ 0.4 & \textbf{+16.6} & \textbf{78.7} & \textbf{+32.7} \\
\rowcolor{RTpihf}\RTin Developed with Luna-low & 56.7 & \textbf{+13.3} & 68.2 $\pm$ 2.2 & \textbf{+22.2} \\
\addlinespace[3pt]
\RTgroup{5}{LiveBench-Math (126 test questions)} \\
\RTin Starting policy & \RTref{58.2} &  & 73.9 &  \\
\rowcolor{RTpihf}\RTin Developed with GPT-4.1-mini & \textbf{63.8} $\pm$ 2.9 & \textbf{+5.6} & \textbf{81.0} & \textbf{+7.1} \\
\rowcolor{RTpihf}\RTin Developed with Luna-low & 60.0 & \textbf{+1.8} & 78.6 $\pm$ 1.0 & \textbf{+4.7} \\
\midrule
\RTpanel{5}{B. Across test rosters} \\
\RTgroup{5}{HoVer, fresh test roster (300 new claims)} \\
\RTin Starting policy & 38.3 &  & 50.3 &  \\
\rowcolor{RTpihf}\RTin Developed with GPT-4.1-mini & \textbf{55.7} & \textbf{+17.3} & \textbf{75.7} & \textbf{+25.3} \\
\bottomrule
\end{tabular}
\endgroup

%% file: manuscript.bbl
\begin{thebibliography}{18}
\providecommand{\natexlab}[1]{#1}
\providecommand{\url}[1]{\texttt{#1}}
\expandafter\ifx\csname urlstyle\endcsname\relax
  \providecommand{\doi}[1]{doi: #1}\else
  \providecommand{\doi}{doi: \begingroup \urlstyle{rm}\Url}\fi

\bibitem[Brown et~al.(2020)Brown, Mann, Ryder, Subbiah, Kaplan, Dhariwal, Neelakantan, Shyam, Sastry, Askell, Agarwal, Herbert-Voss, Krueger, Henighan, Child, Ramesh, Ziegler, Wu, Winter, Hesse, Chen, Sigler, Litwin, Gray, Chess, Clark, Berner, McCandlish, Radford, Sutskever, and Amodei]{brown_language_2020}
Tom~B. Brown, Benjamin Mann, Nick Ryder, Melanie Subbiah, Jared Kaplan, Prafulla Dhariwal, Arvind Neelakantan, Pranav Shyam, Girish Sastry, Amanda Askell, Sandhini Agarwal, Ariel Herbert-Voss, Gretchen Krueger, Tom Henighan, Rewon Child, Aditya Ramesh, Daniel~M. Ziegler, Jeffrey Wu, Clemens Winter, Christopher Hesse, Mark Chen, Eric Sigler, Mateusz Litwin, Scott Gray, Benjamin Chess, Jack Clark, Christopher Berner, Sam McCandlish, Alec Radford, Ilya Sutskever, and Dario Amodei.
\newblock Language {Models} are {Few}-{Shot} {Learners}.
\newblock In \emph{Advances in {Neural} {Information} {Processing} {Systems} 33}, 2020.
\newblock URL \url{https://arxiv.org/abs/2005.14165}.

\bibitem[{Kimi Team} et~al.(2026)]{kimi_team_kimi_2026}
{Kimi Team} et~al.
\newblock Kimi {K3}: {Open} {Frontier} {Intelligence}.
\newblock \emph{arXiv preprint arXiv:2607.24653}, 2026.
\newblock \doi{10.48550/ARXIV.2607.24653}.
\newblock URL \url{https://arxiv.org/abs/2607.24653}.

\bibitem[{DeepSeek-AI} et~al.(2026)]{deepseek-ai_deepseek-v4_2026}
{DeepSeek-AI} et~al.
\newblock {DeepSeek}-{V4}: {Towards} {Highly} {Efficient} {Million}-{Token} {Context} {Intelligence}.
\newblock \emph{arXiv preprint arXiv:2606.19348}, 2026.
\newblock \doi{10.48550/ARXIV.2606.19348}.
\newblock URL \url{https://arxiv.org/abs/2606.19348}.

\bibitem[Marwaha et~al.(2022)Marwaha, Knowles, and Ashley]{marwaha_guide_2022}
Shruti Marwaha, Joshua~W. Knowles, and Euan~A. Ashley.
\newblock A guide for the diagnosis of rare and undiagnosed disease: beyond the exome.
\newblock \emph{Genome Medicine}, 14\penalty0 (1), February 2022.
\newblock ISSN 1756-994X.
\newblock \doi{10.1186/s13073-022-01026-w}.
\newblock URL \url{http://dx.doi.org/10.1186/s13073-022-01026-w}.

\bibitem[Faye et~al.(2024)Faye, Crocione, Anido~de Peña, Bellagambi, Escati~Peñaloza, Hunter, Jensen, Oosterwijk, Schoeters, de~Vicente, Faivre, Wilbur, Le~Cam, and Dubief]{faye_time_2024}
Fatoumata Faye, Claudia Crocione, Roberta Anido~de Peña, Simona Bellagambi, Luciana Escati~Peñaloza, Amy Hunter, Lene Jensen, Cor Oosterwijk, Eva Schoeters, Daniel de~Vicente, Laurence Faivre, Michael Wilbur, Yann Le~Cam, and Jessie Dubief.
\newblock Time to diagnosis and determinants of diagnostic delays of people living with a rare disease: results of a {Rare} {Barometer} retrospective patient survey.
\newblock \emph{European Journal of Human Genetics}, 32\penalty0 (9):\penalty0 1116--1126, May 2024.
\newblock ISSN 1018-4813.
\newblock \doi{10.1038/s41431-024-01604-z}.
\newblock URL \url{http://dx.doi.org/10.1038/s41431-024-01604-z}.

\bibitem[Nguyen et~al.(2026)Nguyen, Gray, Schuler, Byram, Yang, Ma, Xu, Su, Yan, Wei, Wright, Bastarache, Peterson, Li, Ma, {Undiagnosed Diseases Network}, Hamid, Cassini, and Shyr]{nguyen_teaching_2026}
Minh-Ha Nguyen, Erica Gray, Bryce~A. Schuler, Kevin~W. Byram, Chih-Ting Yang, Fan Ma, Hua Xu, Wu-Chen Su, Chao Yan, Wei-Qi Wei, Adam Wright, Lisa Bastarache, Josh Peterson, Lingyao Li, Siyuan Ma, {Undiagnosed Diseases Network}, Rizwan Hamid, Thomas~A. Cassini, and Cathy Shyr.
\newblock Teaching agentic {AI} to generalize expert diagnostic reasoning in rare diseases.
\newblock \emph{arXiv preprint arXiv:2606.16149}, 2026.
\newblock \doi{10.48550/ARXIV.2606.16149}.
\newblock URL \url{https://arxiv.org/abs/2606.16149}.

\bibitem[Radford et~al.(2019)Radford, Wu, Child, Luan, Amodei, and Sutskever]{radford_language_2019}
Alec Radford, Jeffrey Wu, Rewon Child, David Luan, Dario Amodei, and Ilya Sutskever.
\newblock Language {Models} are {Unsupervised} {Multitask} {Learners}.
\newblock Technical report, OpenAI, February 2019.
\newblock URL \url{https://cdn.openai.com/better-language-models/language_models_are_unsupervised_multitask_learners.pdf}.

\bibitem[Agrawal et~al.(2026)Agrawal, Tan, Soylu, Ziems, Khare, Opsahl-Ong, Singhvi, Shandilya, Ryan, Jiang, Potts, Sen, Dimakis, Stoica, Klein, Zaharia, and Khattab]{agrawal_gepa_2026}
Lakshya~A Agrawal, Shangyin Tan, Dilara Soylu, Noah Ziems, Rishi Khare, Krista Opsahl-Ong, Arnav Singhvi, Herumb Shandilya, Michael~J Ryan, Meng Jiang, Christopher Potts, Koushik Sen, Alexandros~G. Dimakis, Ion Stoica, Dan Klein, Matei Zaharia, and Omar Khattab.
\newblock {GEPA}: {Reflective} {Prompt} {Evolution} {Can} {Outperform} {Reinforcement} {Learning}.
\newblock In \emph{The {Fourteenth} {International} {Conference} on {Learning} {Representations}}, 2026.
\newblock URL \url{https://arxiv.org/abs/2507.19457}.

\bibitem[Zhang et~al.(2026)Zhang, Hu, Upasani, Ma, Hong, Kamanuru, Rainton, Wu, Ji, Li, Thakker, Zou, and Olukotun]{zhang_agentic_2026}
Qizheng Zhang, Changran Hu, Shubhangi Upasani, Boyuan Ma, Fenglu Hong, Vamsidhar Kamanuru, Jay Rainton, Chen Wu, Mengmeng Ji, Hanchen Li, Urmish Thakker, James Zou, and Kunle Olukotun.
\newblock Agentic {Context} {Engineering}: {Evolving} {Contexts} for {Self}-{Improving} {Language} {Models}.
\newblock In \emph{The {Fourteenth} {International} {Conference} on {Learning} {Representations}}, 2026.
\newblock URL \url{https://arxiv.org/abs/2510.04618}.

\bibitem[Jin et~al.(2024)Jin, Chen, Zhou, Xu, Cheung, Chen, Summers, Rousseau, Ni, Landsman, Baxter, Al’Aref, Li, Chen, Brejt, Chiang, Peng, and Lu]{jin_hidden_2024}
Qiao Jin, Fangyuan Chen, Yiliang Zhou, Ziyang Xu, Justin~M. Cheung, Robert Chen, Ronald~M. Summers, Justin~F. Rousseau, Peiyun Ni, Marc~J. Landsman, Sally~L. Baxter, Subhi~J. Al’Aref, Yijia Li, Alexander Chen, Josef~A. Brejt, Michael~F. Chiang, Yifan Peng, and Zhiyong Lu.
\newblock Hidden flaws behind expert-level accuracy of multimodal {GPT}-4 vision in medicine.
\newblock \emph{npj Digital Medicine}, 7\penalty0 (1), July 2024.
\newblock ISSN 2398-6352.
\newblock \doi{10.1038/s41746-024-01185-7}.
\newblock URL \url{http://dx.doi.org/10.1038/s41746-024-01185-7}.

\bibitem[Zech et~al.(2018)Zech, Badgeley, Liu, Costa, Titano, and Oermann]{zech_variable_2018}
John~R. Zech, Marcus~A. Badgeley, Manway Liu, Anthony~B. Costa, Joseph~J. Titano, and Eric~Karl Oermann.
\newblock Variable generalization performance of a deep learning model to detect pneumonia in chest radiographs: {A} cross-sectional study.
\newblock \emph{PLOS Medicine}, 15\penalty0 (11):\penalty0 e1002683, November 2018.
\newblock ISSN 1549-1676.
\newblock \doi{10.1371/journal.pmed.1002683}.
\newblock URL \url{http://dx.doi.org/10.1371/journal.pmed.1002683}.

\bibitem[Christiano et~al.(2017)Christiano, Leike, Brown, Martic, Legg, and Amodei]{christiano_deep_2017}
Paul Christiano, Jan Leike, Tom~B. Brown, Miljan Martic, Shane Legg, and Dario Amodei.
\newblock Deep reinforcement learning from human preferences.
\newblock In \emph{Advances in {Neural} {Information} {Processing} {Systems} 30}, 2017.
\newblock URL \url{https://arxiv.org/abs/1706.03741}.

\bibitem[Ouyang et~al.(2022)Ouyang, Wu, Jiang, Almeida, Wainwright, Mishkin, Zhang, Agarwal, Slama, Ray, Schulman, Hilton, Kelton, Miller, Simens, Askell, Welinder, Christiano, Leike, and Lowe]{ouyang_training_2022}
Long Ouyang, Jeff Wu, Xu~Jiang, Diogo Almeida, Carroll~L. Wainwright, Pamela Mishkin, Chong Zhang, Sandhini Agarwal, Katarina Slama, Alex Ray, John Schulman, Jacob Hilton, Fraser Kelton, Luke Miller, Maddie Simens, Amanda Askell, Peter Welinder, Paul Christiano, Jan Leike, and Ryan Lowe.
\newblock Training language models to follow instructions with human feedback.
\newblock In \emph{Advances in {Neural} {Information} {Processing} {Systems} 35}, 2022.
\newblock \doi{10.52202/068431-2011}.
\newblock URL \url{https://arxiv.org/abs/2203.02155}.

\bibitem[Kazemi et~al.(2025)Kazemi, Fatemi, Bansal, Palowitch, Anastasiou, Mehta, Jain, Aglietti, Jindal, Chen, Dikkala, Tyen, Liu, Shalit, Chiappa, Olszewska, Tay, Tran, Le, and Firat]{kazemi_big-bench_2025}
Mehran Kazemi, Bahare Fatemi, Hritik Bansal, John Palowitch, Chrysovalantis Anastasiou, Sanket~Vaibhav Mehta, Lalit~K. Jain, Virginia Aglietti, Disha Jindal, Peter Chen, Nishanth Dikkala, Gladys Tyen, Xin Liu, Uri Shalit, Silvia Chiappa, Kate Olszewska, Yi~Tay, Vinh~Q. Tran, Quoc~V. Le, and Orhan Firat.
\newblock {BIG}-{Bench} {Extra} {Hard}.
\newblock In \emph{Proceedings of the 63rd {Annual} {Meeting} of the {Association} for {Computational} {Linguistics} ({Volume} 1: {Long} {Papers})}, pp.\  26473--26501. Association for Computational Linguistics, 2025.
\newblock \doi{10.18653/v1/2025.acl-long.1285}.
\newblock URL \url{https://arxiv.org/abs/2502.19187}.

\bibitem[Jiang et~al.(2020)Jiang, Bordia, Zhong, Dognin, Singh, and Bansal]{jiang_hover_2020}
Yichen Jiang, Shikha Bordia, Zheng Zhong, Charles Dognin, Maneesh Singh, and Mohit Bansal.
\newblock {HoVer}: {A} {Dataset} for {Many}-{Hop} {Fact} {Extraction} {And} {Claim} {Verification}.
\newblock In \emph{Findings of the {Association} for {Computational} {Linguistics}: {EMNLP} 2020}, pp.\  3441--3460, 2020.
\newblock \doi{10.18653/v1/2020.findings-emnlp.309}.
\newblock URL \url{http://dx.doi.org/10.18653/v1/2020.findings-emnlp.309}.

\bibitem[White et~al.(2025)White, Dooley, Roberts, Pal, Feuer, Jain, Shwartz-Ziv, Jain, Saifullah, Dey, Shubh-Agrawal, Sandha, Naidu, Hegde, LeCun, Goldstein, Neiswanger, and Goldblum]{white_livebench_2025}
Colin White, Samuel Dooley, Manley Roberts, Arka Pal, Ben Feuer, Siddhartha Jain, Ravid Shwartz-Ziv, Neel Jain, Khalid Saifullah, Sreemanti Dey, Shubh-Agrawal, Sandeep~Singh Sandha, Siddartha Naidu, Chinmay Hegde, Yann LeCun, Tom Goldstein, Willie Neiswanger, and Micah Goldblum.
\newblock {LiveBench}: {A} {Challenging}, {Contamination}-{Limited} {LLM} {Benchmark}.
\newblock In \emph{The {Thirteenth} {International} {Conference} on {Learning} {Representations}}, 2025.
\newblock URL \url{https://arxiv.org/abs/2406.19314}.

\bibitem[Sutton \& Barto(2018)Sutton and Barto]{sutton_reinforcement_2018}
Richard~S. Sutton and Andrew~G. Barto.
\newblock \emph{Reinforcement {Learning}: {An} {Introduction}}.
\newblock MIT Press, second edition, November 2018.
\newblock ISBN 9780262039246.

\bibitem[Chan et~al.(2022)Chan, Santoro, Lampinen, Wang, Singh, Richemond, McClelland, and Hill]{chan_data_2022}
Stephanie C.~Y. Chan, Adam Santoro, Andrew~K. Lampinen, Jane~X. Wang, Aaditya Singh, Pierre~H. Richemond, Jay McClelland, and Felix Hill.
\newblock Data {Distributional} {Properties} {Drive} {Emergent} {In}-{Context} {Learning} in {Transformers}.
\newblock In \emph{Advances in {Neural} {Information} {Processing} {Systems} 35}, pp.\  18878--18891, 2022.
\newblock \doi{10.52202/068431-1371}.
\newblock URL \url{https://arxiv.org/abs/2205.05055}.

\end{thebibliography}
